%% file: main.tex
\documentclass[10pt,twocolumn,letterpaper]{article}

\usepackage{cvpr}

\usepackage{epsfig}
\usepackage{graphicx}
\usepackage{amsmath}
\usepackage{dsfont}
\usepackage{fixmath}
\usepackage{amssymb}
\usepackage{float} 
\usepackage{subfigure}
\usepackage{color}
\usepackage{booktabs}
\input{macro}

\graphicspath{ {figure/} }

\newcommand{\ra}[1]{\renewcommand{\arraystretch}{#1}}
\newcolumntype{x}[1]{>{\centering\arraybackslash}p{#1pt}}
\newcommand{\tablestyle}[2]{\setlength{\tabcolsep}{#1}\renewcommand{\arraystretch}{#2}\centering\footnotesize}
\newlength\savewidth\newcommand\shline{\noalign{\global\savewidth\arrayrulewidth
		\global\arrayrulewidth 1pt}\hline\noalign{\global\arrayrulewidth\savewidth}}

\makeatletter\renewcommand\paragraph{\@startsection{paragraph}{4}{\z@}
	{.5em \@plus1ex \@minus.2ex}{-.5em}{\normalfont\normalsize\bfseries}}\makeatother

\usepackage[pagebackref=true,breaklinks=true,letterpaper=true,colorlinks,bookmarks=false]{hyperref}

 \cvprfinalcopy %

\def\cvprPaperID{161} %

\ifcvprfinal\fi
\begin{document}

\title{Context Prior for Scene Segmentation}

\author{
    Changqian Yu$^{1,2}$
    ~
    Jingbo Wang$^3$
    ~
    {Changxin Gao$^1$}\thanks{Corresponding author. 
    Part of the work was done when C. Yu was visiting The University of Adelaide.}
    ~
    Gang Yu$^4$
    ~
    Chunhua Shen$^2$
    ~ 
    Nong Sang$^1$\\
    \normalsize
    $^1$Key Laboratory of Image Processing and Intelligent Control,\\
    \normalsize
    School of Artificial Intelligence and Automation, Huazhong University of Science and Technology\\
    \normalsize
    $^2$The University of Adelaide, Australia
    ~~
    $^3$The Chinese University of Hong Kong
    ~~
    $^4$Tencent\\
{\small \tt \{changqian\_yu,cgao,nsang\}@hust.edu.cn%
} \\
}

\maketitle
\thispagestyle{empty}

\begin{abstract}
Recent works have widely explored the contextual dependencies to 
achieve 
more accurate segmentation results.
However, most approaches rarely distinguish different types of contextual dependencies, which may pollute the scene understanding.
In this work, we directly supervise the feature aggregation to distinguish the intra-class and inter-class context clearly.
Specifically, we develop a Context Prior with the supervision of the Affinity Loss.
Given an input image and corresponding ground truth, Affinity Loss constructs an ideal affinity map to supervise the learning of Context Prior.
The learned Context Prior extracts the pixels belonging to the same category, while the reversed prior focuses on the pixels of different classes. 
Embedded into a conventional deep CNN, the proposed Context Prior Layer can selectively capture the intra-class and inter-class contextual dependencies, %
leading to robust feature representation.
To validate 
the
effectiveness, we design an effective Context Prior Network (CPNet).
Extensive quantitative and qualitative evaluations demonstrate that the proposed model performs favorably against 
state-of-the-art semantic segmentation approaches.
More specifically, our algorithm achieves $46.3\%$ mIoU on ADE20K, $53.9\%$ mIoU on PASCAL-Context, and $81.3\%$ mIoU on Cityscapes.
Code is available at \url{https://git.io/ContextPrior}.
\end{abstract}

\section{Introduction}
\label{sec:intro}
Scene segmentation is a long-standing and challenging problem in computer vision with 
many downstream
applications \eg augmented reality, autonomous driving~\cite{Cityscapes, Kitti}, human-machine interaction, and video 
content analysis.
The goal is to assign each pixel with a category label, which provides 
comprehensive scene understanding.

\begin{figure}[t]
\footnotesize
\centering
\renewcommand{\tabcolsep}{1pt} %
\ra{1} %
\begin{center}
\begin{tabular}{ccc}
\includegraphics[width=0.33\linewidth]{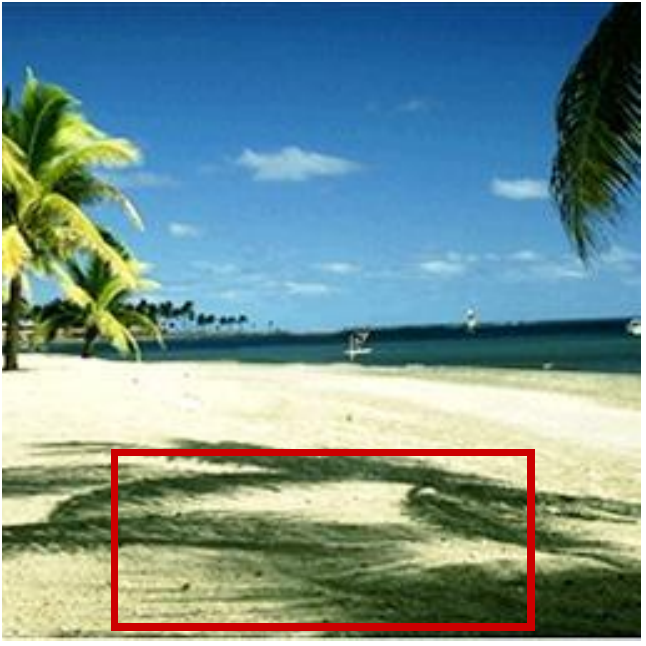} &
\includegraphics[width=0.33\linewidth]{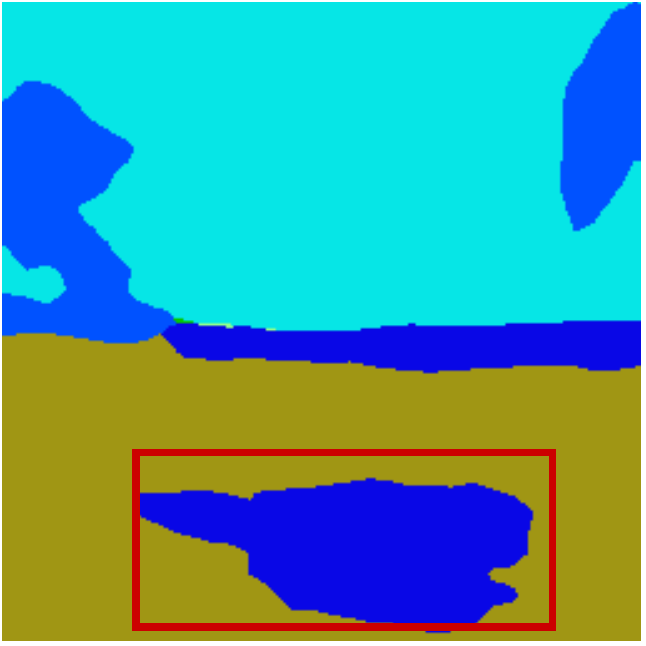} &
\includegraphics[width=0.33\linewidth]{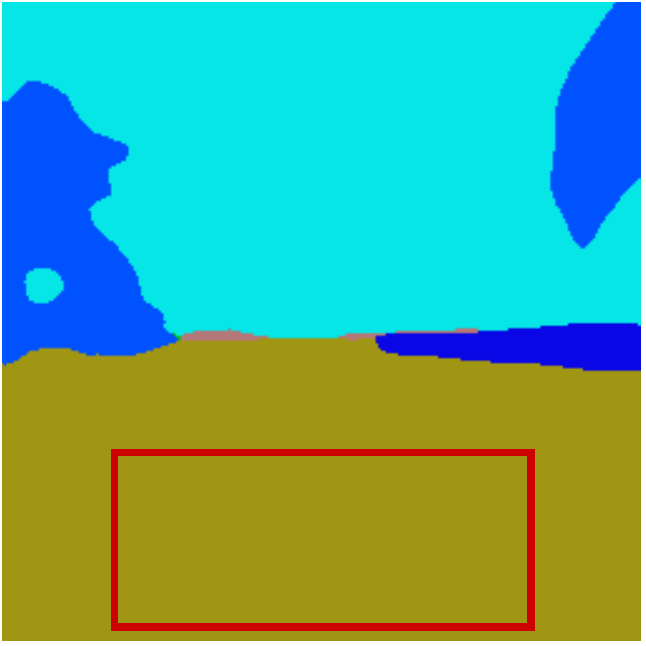} \\
(a) Input Image &
(b) Pyramid Method &
(c) CPNet \\
\includegraphics[width=0.33\linewidth]{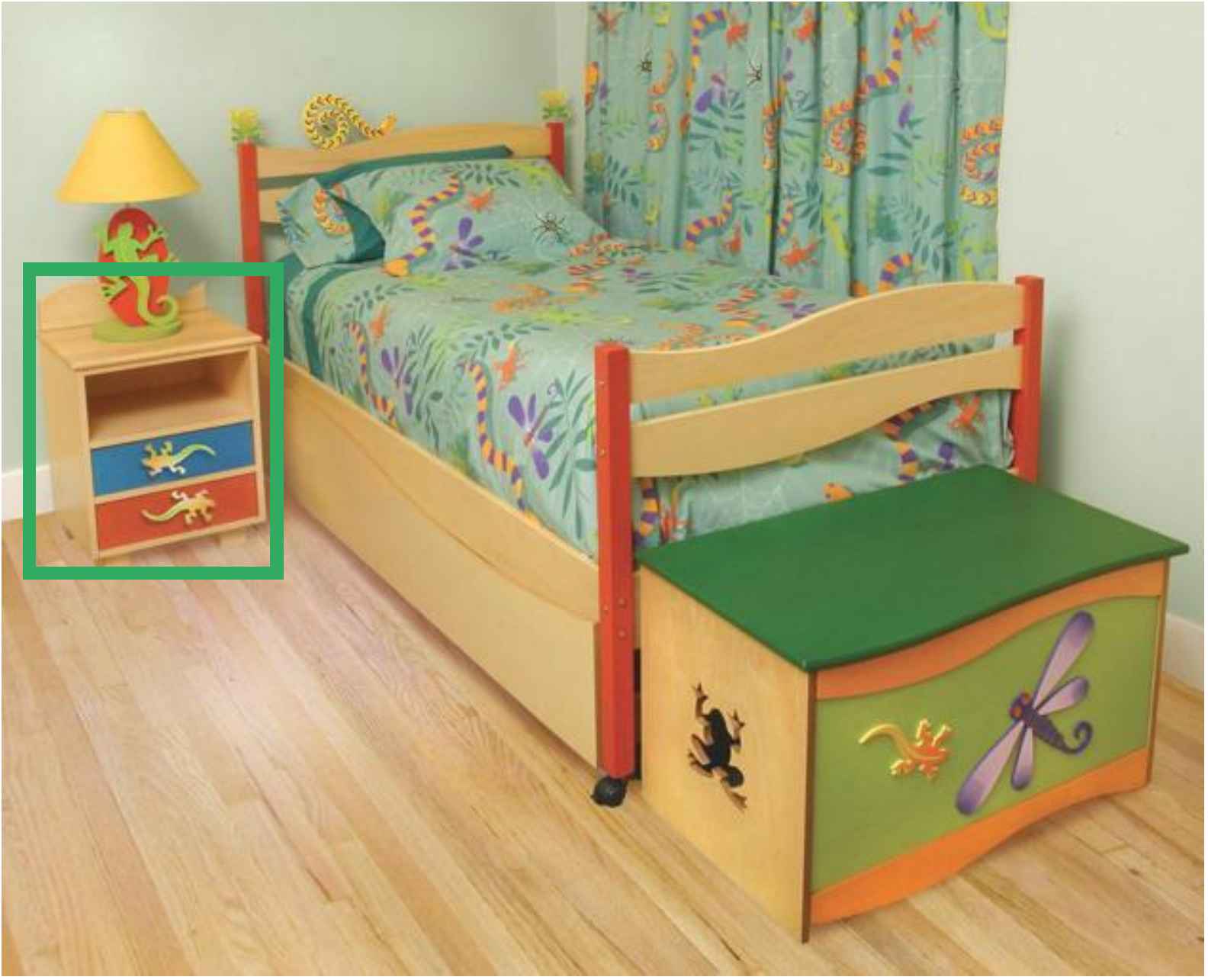} &
\includegraphics[width=0.33\linewidth]{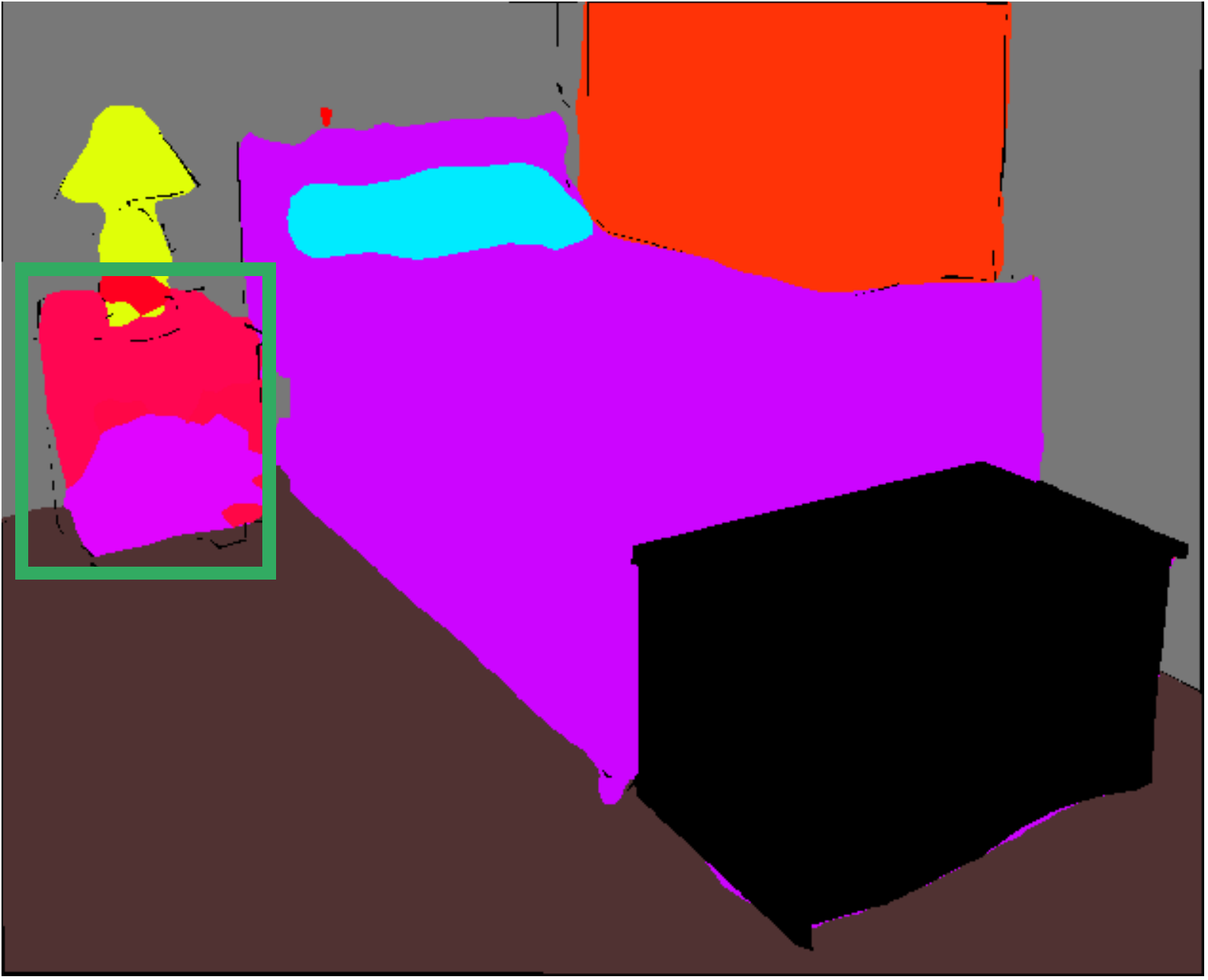} &
\includegraphics[width=0.33\linewidth]{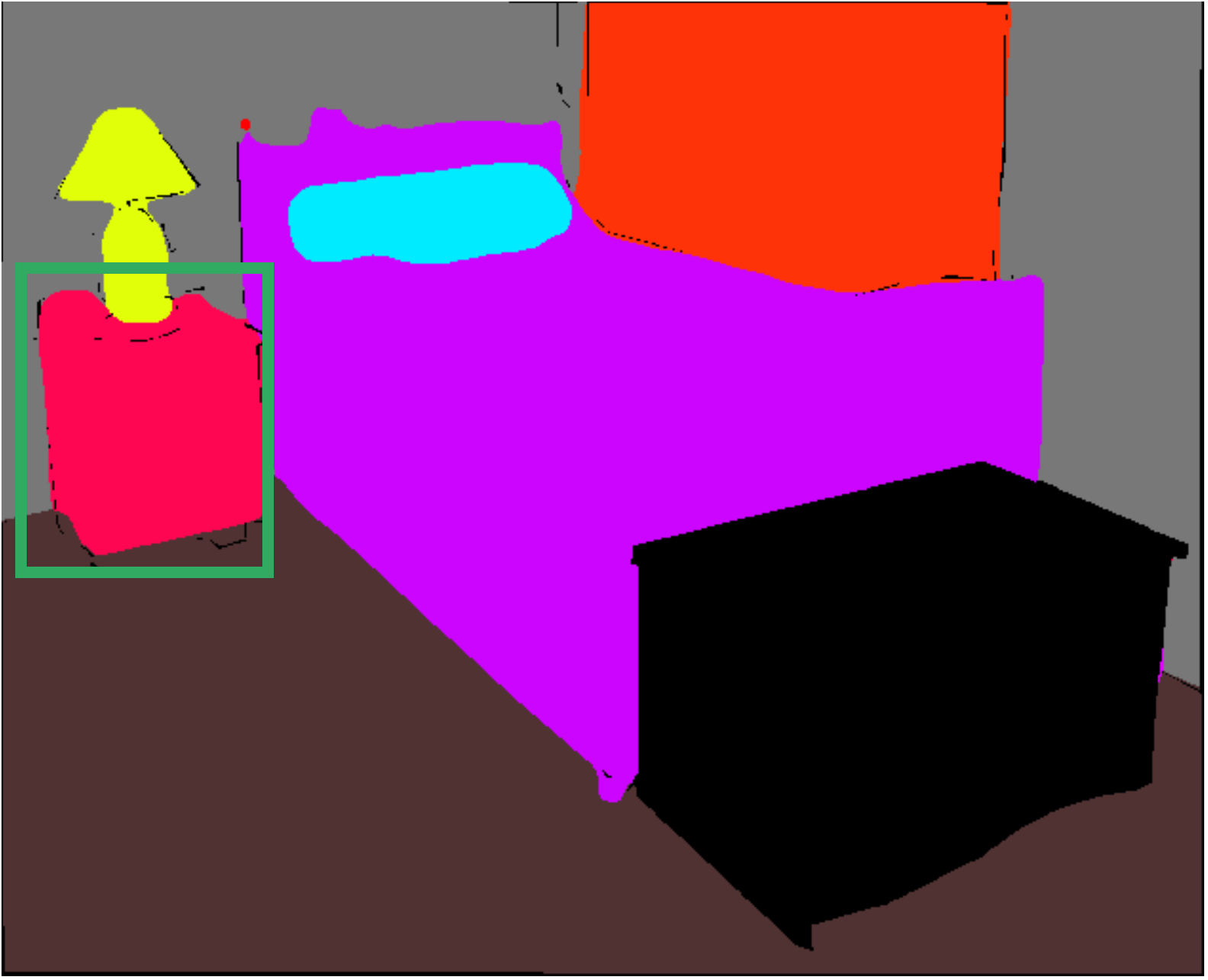} \\
(d) Input Image &
(e) Attention Method &
(f) CPNet \\
\end{tabular}
\end{center}
\caption{\textbf{Hard examples in scene segmentation.}
In the first row, the central part of the \textit{sand} in the red box is misclassified as the \textit{sea}, because the shadow part has a similar appearance with the \textit{sea}.
With the pyramid-based aggregation method~\cite{Chen-Arxiv-Deeplabv3-2017}, aggregation of the confused spatial information may lead to undesirable prediction as visualized in (b).
In the second row, the \textit{table} in the green box has a similar appearance to the bottom part of the bed.
The attention-based method~\cite{Zhao-ECCV-PSANet-2018} 
fails to
effectively distinguish the confused spatial information without 
prior knowledge, leading to less correct prediction as shown in (e).
In the proposed CPNet, we aggregate the contextual dependencies with clear distinguishment.
Notably, the Context Prior models the intra-class and inter-class relationships as a context prior knowledge to capture the intra-class and inter-class contextual dependencies.
}
\label{fig:fig1}
\end{figure}

Benefiting
from the effective feature representation of the Fully Convolutional Network (FCN), 
a few
approaches have obtained promising performance.
However, limited by the structure of convolutional layers, the FCN provides insufficient contextual information, 
leaving room for improvement.  
Therefore, various methods~\cite{Chen-ICLR-Deeplabv2-2016, Chen-Arxiv-Deeplabv3-2017, Chen-ECCV-Deeplabv3p-2018, Peng-CVPR-Largekernl-2017, Zhao-CVPR-PSPNet-2017, Yu-CVPR-DFN-2018, Zhang-CVPR-EncNet-2018, Wang-TPAMI-HRNet-2019, Hung-ICCV-GCPNet-2017} explore the contextual dependencies to obtain more accurate segmentation results.
There are mainly two paths to aggregate the contextual information: 1) Pyramid-based aggregation method.
Several methods~\cite{Zhao-CVPR-PSPNet-2017, Chen-ICLR-Deeplabv2-2016, Chen-Arxiv-Deeplabv3-2017, Chen-ECCV-Deeplabv3p-2018} adopt pyramid-based modules or global pooling to aggregate regional or global contextual details regularly.
However, they capture the homogeneous contextual relationship, ignoring the contextual dependencies of different categories, as shown in Figure~\ref{fig:fig1}(b).
When there are confused categories in the scene, these methods may result in a less reliable context.
2) Attention-based aggregation method.
Recent attention-based methods learn channel attention~\cite{Zhang-CVPR-EncNet-2018, Yu-CVPR-DFN-2018}, spatial attention~\cite{Li-BMVC-PAN-2018}, or point-wise attention~\cite{Zhao-ECCV-PSANet-2018, Fu-CVPR-DANet-2019, Yuan-Arxiv-OCNet-2018} to aggregate the heterogeneous contextual information selectively.
Nevertheless, due to the lack of explicit regularization, the relationship description of the attention mechanism is less clear.
Therefore, it may select undesirable contextual dependencies, as visualized in Figure~\ref{fig:fig1}(e).
Overall, both paths aggregate contextual information without explicit distinction, causing a mixture of different contextual relationships.

We notice that the identified contextual dependencies help the network understand the scene.
The correlation of the same category (intra-class context) and the difference between the different classes (inter-class context) make the feature representation more robust and reduce the search space of possible categories.
Therefore, we model the contextual relationships among categories as prior knowledge to obtain more accurate prediction, which is of great importance to the scene segmentation.

In this paper, we construct a \textbf{Context Prior} to model the intra-class and inter-class dependencies as the prior knowledge.
We formulate the context prior as a binary classifier to distinguish which pixels belong to the same category for the current pixel, while the reversed prior can focus on the pixels of different classes. 
Specifically, we first 
use 
a fully convolutional network to generate the feature map and the corresponding prior map.
For each pixel in the feature map, the prior map can selectively highlight other pixels belonging to the same category to aggregate the intra-class context, while the reversed prior can aggregate the inter-class context.
To embed the prior into the network, we develop a \textbf{Context Prior Layer} incorporating an \textbf{Affinity Loss}, which directly supervises the learning of the prior.
Meanwhile, Context Prior also requires spatial information to reason the relationships.
To this end, we design an \textbf{Aggregation Module}, which adopts the fully separable convolution (separate on both the spatial and depth dimensions)~\cite{Peng-CVPR-Largekernl-2017, Chollet-CVPR-Xception-2017, Zhang-CVPR-Shufflenet-2018, Ma-ECCV-Shufflenetv2-2018} to efficiently aggregate spatial information.

To demonstrate the effectiveness of the proposed Context Prior, we design a simple fully convolutional network called \textbf{Context Prior Network (CPNet)}.
Based on the output features of the backbone network~\cite{Chen-ICLR-Deeplabv2-2016, Chen-Arxiv-Deeplabv3-2017, Wang-WACV-DUC-2018}, the Context Prior Layer 
uses
the Aggregation Module to aggregate the spatial information to generate a Context Prior Map.
With the supervision of Affinity Loss, the Context Prior Map can capture intra-class context and inter-class context to refine the prediction.
Extensive evaluations demonstrate that the proposed method performs favorably against 
several recent
\emph{state-of-the-art} semantic segmentation approaches.

The main contributions of this work are summarized as follows.
\begin{compactitem}
\item We construct a Context Prior with supervision of an Affinity Loss embedded in a Context Prior Layer to capture the intra-class and inter-class contextual dependencies explicitly.
\item We design an effective Context Prior Network (CPNet) for scene segmentation, which contains a backbone network and a Context Prior Layer.
\item We demonstrate the proposed method performs favorably against \emph{state-of-the-art} approaches on the benchmarks of ADE20K, Pascal-Context, and Cityscapes. 
More specifically, our single model achieves $46.3\%$ on the ADE20K validation set, $53.9\%$ on the PASCAL-Context validation set and $81.3\%$ on the Cityscapes test set.
\end{compactitem}

\section{Related Work}
\paragraph{Context Aggregation.}
In recent years, various methods have explored contextual information, which is crucial to scene understanding~\cite{Chen-ICLR-Deeplabv2-2016, Chen-ECCV-Deeplabv3p-2018, Peng-CVPR-Largekernl-2017, Zhao-CVPR-PSPNet-2017, Yu-CVPR-DFN-2018, Zhang-CVPR-EncNet-2018, Yuan-Arxiv-OCNet-2018, Hung-ICCV-GCPNet-2017, Liu-CVPR-OANet-2019, Yin-ICCV-VN-2019}. 
There are mainly two paths to capture contextual dependencies. 
1) PSPNet~\cite{Zhao-CVPR-PSPNet-2017} adopts the pyramid pooling module to partition the feature map into different scale regions. 
It averages the pixels of each area as the local context of each pixel in this region. 
Meanwhile, Deeplab~\cite{Chen-ICLR-Deeplabv2-2016, Chen-Arxiv-Deeplabv3-2017, Chen-ECCV-Deeplabv3p-2018} methods employ atrous spatial pyramid pooling to sample the different range of pixels as the local context. 
2) DANet~\cite{Fu-CVPR-DANet-2019}, OCNet~\cite{Yuan-Arxiv-OCNet-2018}, and CCNet~\cite{Huang-ICCV-CCNet-2018} take advantage of the self-similarity manner~\cite{Wang-CVPR-Nonlocal-2018} to aggregate long-range spatial information. 
Besides, EncNet~\cite{Zhang-CVPR-EncNet-2018}, DFN~\cite{Yu-CVPR-DFN-2018}, and ParseNet~\cite{Liu-ARXIV-ParseNet-2016} use  global pooling to harvest the global context.

Despite the success of these attention mechanisms, they maybe capture undesirable contextual dependencies without explicitly distinguishing the difference of different contextual relationships.
Therefore, in the proposed approach, we explicitly regularize the model to obtain the intra-class and inter-class contextual dependencies.

\begin{figure*}[t]
\centering
\includegraphics[width=0.96\linewidth]{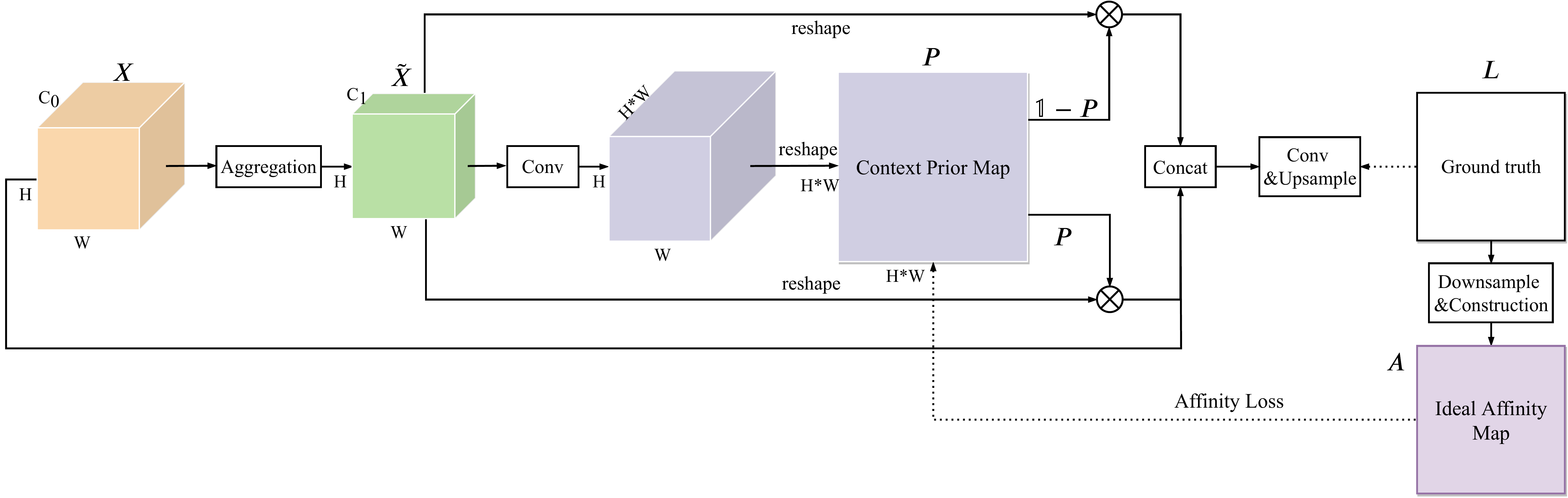}
\caption{\textbf{Overview of the proposed Context Prior Layer}.
The Context Prior Layer contains an Aggregation Module and a Context Prior Map supervised by Affinity Loss.
With the extracted input features, the Aggregation Module aggregates the spatial information to reason the contextual relationship.
We generate a point-wise Context Prior Map with the supervision of an Affinity Loss.
The Affinity Loss constructs an Ideal Affinity Map which indicates the pixels of the same category to supervise the learning of the Context Prior Map.
Based on the Context Prior Map, we can obtain the intra-prior ($\mathbold{P}$) and inter-prior ($\mathds{1} - \mathbold{P}$).
The original feature map is reshaped to $N \times C_1$ size, where $N = H \times W$.
We conduct matrix multiplication on the reshaped feature map with $\mathbold{P}$ and $(\mathds{1} - \mathbold{P})$ to capture the intra-class and inter-class context.
Finally, we feed the representation of the Context Prior Layer into the last convolutional layer to generate a per-pixel prediction.
(Notation: \textit{Aggregation} Aggregation Module, \textit{Conv} convolutional layer, $\bigotimes$ matrix multiplication, $\mathbold{P}$ Context Prior Map, \textit{Concat} concatenate operation).
}
\label{fig:network-structure}
\end{figure*}

\paragraph{Attention Mechanism.}
Recent years have witnessed the broad application of the attention mechanism.
It can be used for various tasks such as machine translation~\cite{Ashish-NIPS-Transfomer-2017}, image/action recognition~\cite{Wang-CVPR-Nonlocal-2018, Chen-NIPS-A2Net-2018, Hu-CVPR-SEnet-2017}, object detection~\cite{Hu-CVPR-RelationNet-2018} and semantic segmentation~\cite{Yu-CVPR-DFN-2018, Zhang-CVPR-EncNet-2018, Yu-ECCV-BiSeNet-2018, Zhao-ECCV-PSANet-2018, Fu-CVPR-DANet-2019, Yuan-Arxiv-OCNet-2018}.

For the semantic segmentation task, \cite{Chen-CVPR-AttentionScale-2016} learns an attention mechanism to weight the multi-scale features softly.
Inspired by SENet~\cite{Hu-CVPR-SEnet-2017}, some methods such as EncNet~\cite{Zhang-CVPR-EncNet-2018}, DFN~\cite{Yu-CVPR-DFN-2018}, and BiSeNet~\cite{Yu-ECCV-BiSeNet-2018} adopt the channel attention to select the desired feature map.
Following ~\cite{Ashish-NIPS-Transfomer-2017, Wang-CVPR-Nonlocal-2018}, DANet~\cite{Fu-CVPR-DANet-2019} and OCNet~\cite{Yuan-Arxiv-OCNet-2018} use the self-attention to capture the long-range dependency, while PSANet~\cite{Zhao-ECCV-PSANet-2018} adaptively learns point-wise attention to harvest the long-range information.
However, these effective methods lack the explicit regularization, maybe leading to an undesirable context aggregation.
Therefore, in our work, we propose a Context Prior 
embedded 
in the Context Prior Layer with an explicit Affinity Loss to supervise the learning process.

\section{Context Prior}
\label{sec:context-prior}
Contextual dependencies play a crucial role in scene understanding, which is widely explored in various methods~\cite{Zhao-CVPR-PSPNet-2017, Peng-CVPR-Largekernl-2017, Liu-ARXIV-ParseNet-2016, Zhang-CVPR-EncNet-2018, Chen-Arxiv-Deeplabv3-2017, Yu-CVPR-DFN-2018}.
However, these methods aggregate different contextual dependencies as a mixture. 
As discussed in Section~\ref{sec:intro}, the clear distinguished contextual relationships are 
desirable 
to the scene understanding.

In our study, we propose a Context Prior to model the relationships between pixels of the same category (intra-context) and pixels of the different categories (inter-context).
Based on the Context Prior, we propose a Context Prior Network,
incorporating 
a Context Prior Layer with the supervision of an Affinity Loss, as shown in Figure~\ref{fig:network-structure}.
In this section, we first introduce the Affinity Loss, which supervises the layer to learn a Context Prior Map.
Next, we demonstrate the Context Prior Layer, which uses the learned Context Prior Map to aggregate the intra-context and inter-context for each pixel.
The Aggregation Module is designed to aggregate the spatial information for reasoning.
Finally, we elaborate on our complete network structure.

\subsection{Affinity Loss}
In the scene segmentation task, for each image, we have one ground truth, which assigns a semantic category for each pixel.
It is hard for the network to model the contextual information from isolated pixels.
To explicitly regularize the network to model the relationship between categories, we introduce an Affinity Loss.
For each pixel in the image, this loss forces the network to consider the pixels of the same category (intra-context) and the pixels among the different categories (inter-context).

\begin{figure}[t]
\centering
\includegraphics[width=0.96\linewidth]{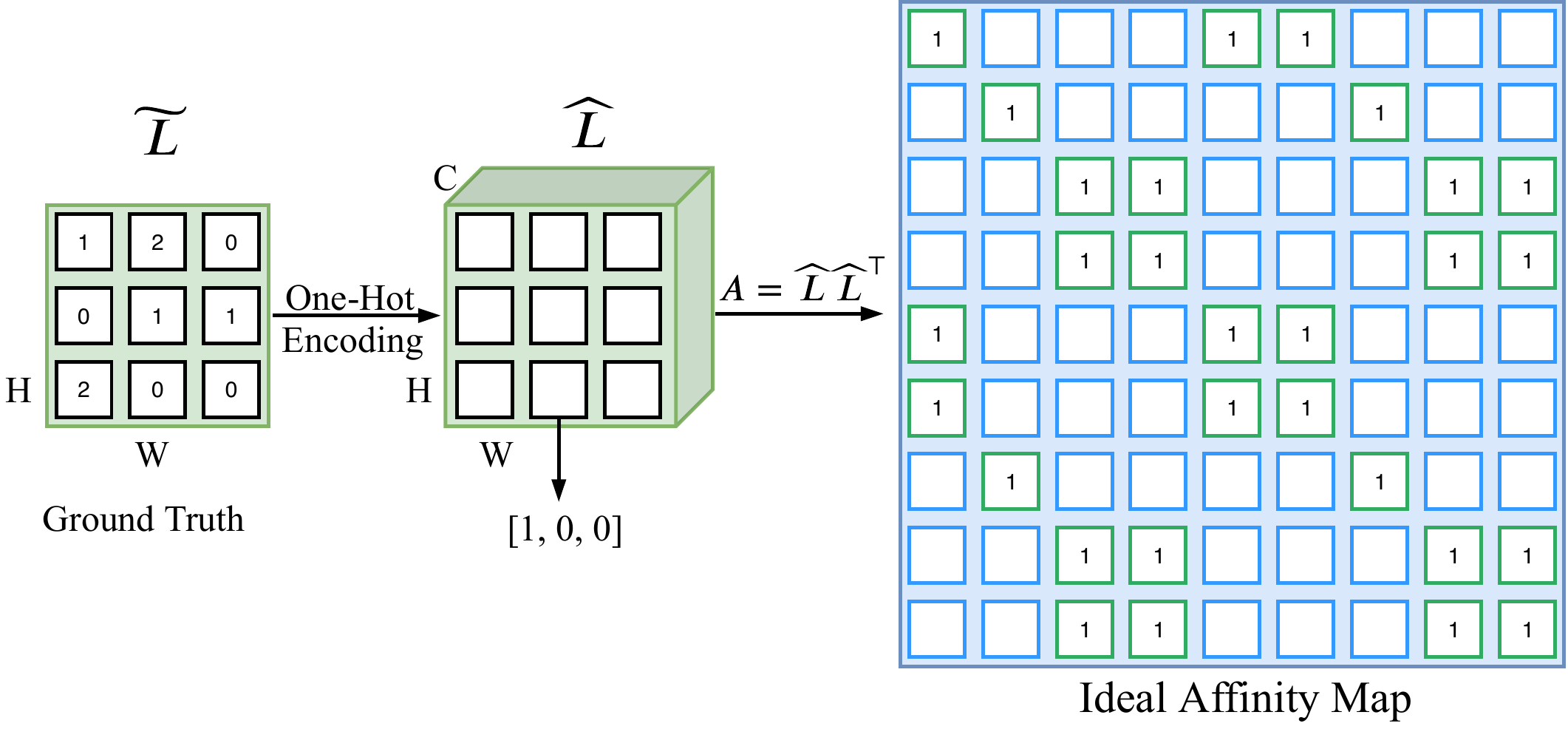}
\caption{\textbf{Illustration of the construction of the Ideal Affinity Map}.
The downsampled ground truth $\mathbold{\widetilde{L}}$ is first encoded with the one-hot encoding.
The size of the ground truth $\mathbold{\widehat{L}}$ becomes $H \times W \times C$, where $C$ is the number of the classes.
Each vector in $\mathbold{\widehat{L}}$ is composed of a single high value (1) and all the others low (0).
We conduct $\mathbold{A} = \mathbold{\widehat{L}} \mathbold{\widehat{L}^\top}$ to generate the Ideal Affinity Map.
In this map, the green box and blue box represent 1 and 0, respectively.
}
\label{fig:prior-affinity}
\end{figure}

Given a ground truth for an input, we can know the ``context prior'' of each pixel (\ie which pixels belong to the same category and which pixels do not).
Therefore, we can learn a Context Prior to guiding the network according to the ground truth.
To this end, we first construct an Ideal Affinity Map from the ground truth as the supervision.
Given 
an
input image $\mathbold{I}$ and the ground truth $\mathbold{L}$, we feed the input image $\mathbold{I}$ to the network, obtaining a feature map $\mathbold{X}$ of size $H \times W$.
As shown in Figure~\ref{fig:prior-affinity}, we first down-sample the ground truth $\mathbold{L}$ into the same size of the feature map $\mathbold{X}$, yielding a smaller ground truth $\mathbold{\widetilde{L}}$.
We use a one-of-K scheme (one-hot encoding) to encode each categorical integer label in the ground truth $\mathbold{\widetilde{L}}$, leading to a matrix $\mathbold{\widehat{L}}$ of $H \times W \times C$ size, where $C$ is the number of classes.
Next, we reshape the encoded ground truth to $N \times C$ size, in which $N = H \times W$.
Finally, we conduct the matrix multiplication: $\mathbold{A} = \mathbold{\widehat{L}} \mathbold{\widehat{L}^\top}$.
$\mathbold{A}$ is our desired Ideal Affinity Map with size $N \times N$, which encodes which pixels belong to the same category.
We 
employ
the Ideal Affinity Map to supervise the learning of Context Prior Map.

For each pixel in the prior map, it is a binary classification problem.
A conventional method for addressing this problem is to use the binary cross entropy loss.
Given the predicted Prior Map $\mathbold{P}$ of size $N \times N$, where $\lbrace p_n \in \mathbold{P}, n \in \lbrack 1, N^2 \rbrack  \rbrace$ and the reference Ideal Affinity Map $\mathbold{A}$, where $\lbrace a_n \in \mathbold{A}, n \in \lbrack 1, N^2 \rbrack  \rbrace$, the binary cross entropy loss can be denoted as:
\begin{equation}\label{eq:bce}
\mathcal{L}_{u} = - \frac{1}{N^2}\sum^{N^2}_{n=1}{(a_n\log{p_n}+(1-a_n)\log{(1-p_n)})}.
\end{equation}
However, such a unary loss 
only
considers the isolated pixel in the prior map ignoring the semantic correlation with other pixels.
The pixels of each row of the Prior Map $\mathbold{P}$ is corresponding to the pixels of the feature map $\mathbold{X}$.
We can divide them into intra-class pixels and inter-class pixels, the relationships of which are helpful to reason the semantic correlation and scene structure.
Therefore, we can consider the intra-class pixels and inter-class pixels as two wholes to encode the relationships respectively.
To this end, we devise the global term based on the binary cross entropy loss:
\begin{align}
\label{eq:global-term}
\mathcal{T}^{p}_{j} &= \log{\frac{\sum^{N}_{i=1}{a_{ij} p_{ij}}}{\sum^{N}_{i=1}{p_{ij}}}},\\
\mathcal{T}^{r}_{j} &= \log{\frac{\sum^{N}_{i=1}{a_{ij} p_{ij}}}{\sum^{N}_{i=1}{a_{ij}}}},\\
\mathcal{T}^{s}_{j} &= \log{\frac{\sum^{N}_{i=1}{(1-a_{ij}) (1-p_{ij})}}{\sum^{N}_{i=1}{(1-a_{ij})}}},\\
\mathcal{L}_g &= - \frac{1}{N} \sum^{N}_{j=1}{(\mathcal{T}^{p}_{j} + \mathcal{T}^{r}_{j} + \mathcal{T}^{s}_{j})},
\end{align}
where $\mathcal{T}^{p}_{j}$, $\mathcal{T}^{r}_{j}$, and $\mathcal{T}^{s}_{j}$ represent the intra-class predictive value (precision), true intra-class rate (recall), and true inter-class rate (specificity) at $j^{th}$ row of $\mathbold{P}$, respectively.
Finally, based on both the unary term and global term, the complete Affinity Loss can be denoted as follows:
\begin{equation}
	\mathcal{L}_{p} = \lambda_u \mathcal{L}_{u} + \lambda_g \mathcal{L}_{g},
\end{equation}
where $\mathcal{L}_{p}$, $\mathcal{L}_{u}$, and $\mathcal{L}_{g}$ represent the affinity loss, unary loss (binary cross entropy loss), and global loss functions, respectively.
In addition, $\lambda_u$ and $\lambda_g$ are the balance weights for the unary loss and global loss, respectively.
We empirically set the weights as: $\lambda_u=1$ and $\lambda_g=1$.

\begin{figure}[t]
\footnotesize
\centering
\renewcommand{\tabcolsep}{1pt} %
\ra{1} %
\begin{center}
\begin{tabular}{cc}
\includegraphics[width=0.4\linewidth]{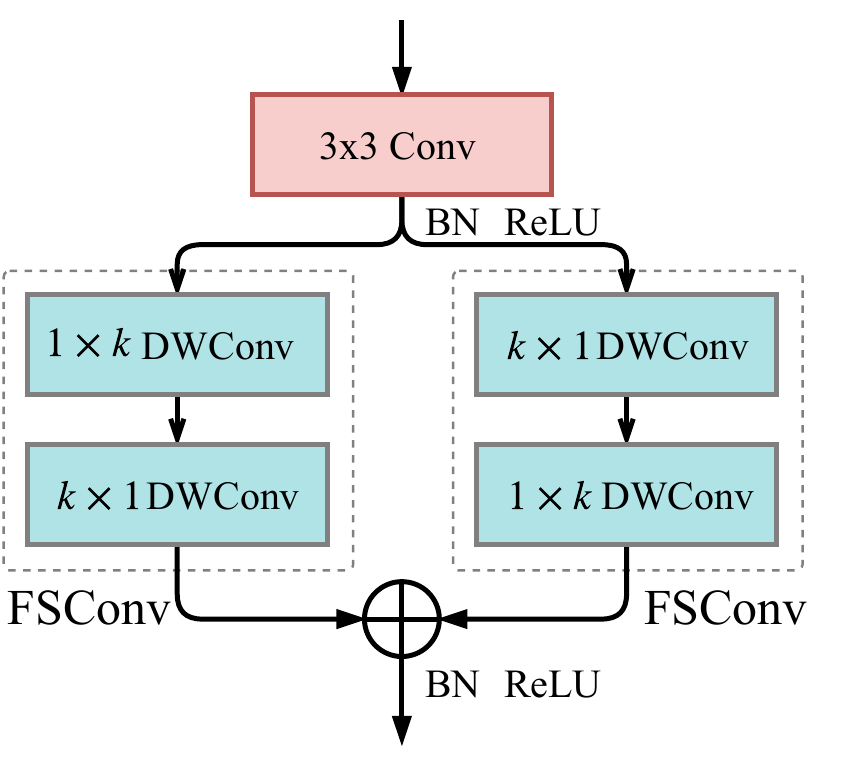} &
\includegraphics[width=0.48\linewidth]{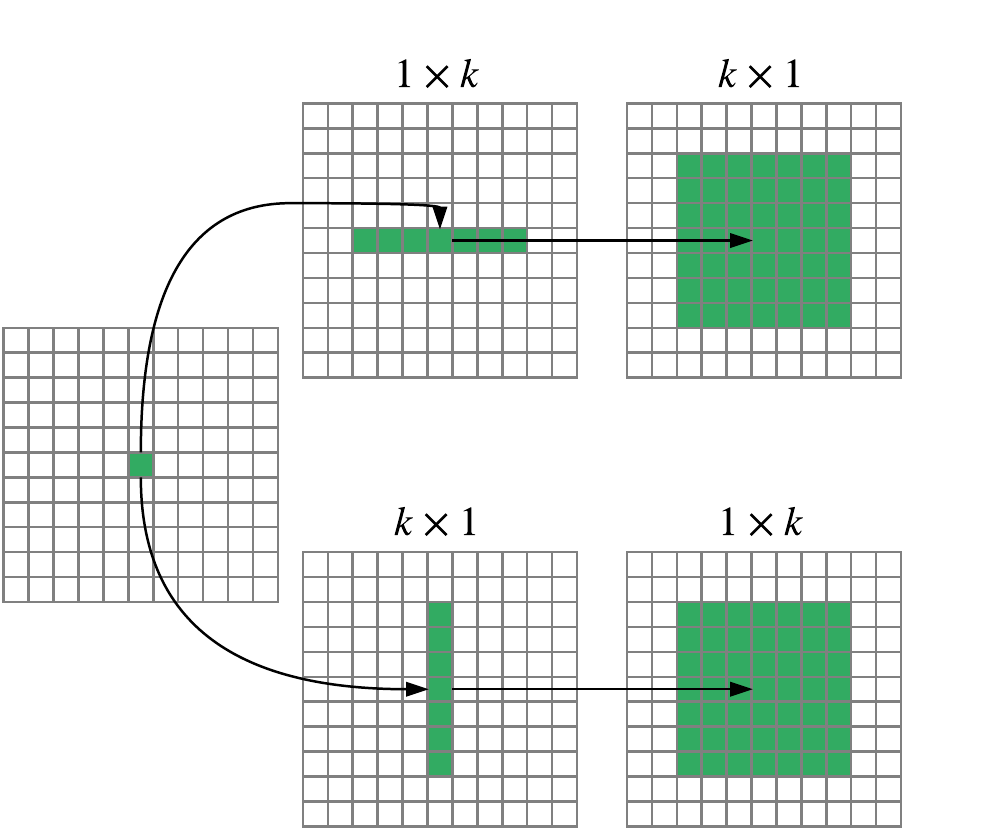} \\
(a) Aggregation Module &
(b) Receptive filed of Aggregation Module \\
\end{tabular}
\end{center}
\caption{\textbf{Illustration of the Aggregation Module and its receptive field}.
(a) We use two asymmetric fully separable convolutions to aggregate the spatial information, the output of which has the same channels with the input features. 
(b) The Aggregation Module has the same size of receptive field with the standard convolution.
However, our Aggregation Module leads to less computation.
(Notation: \textit{Conv} standard convolution, \textit{DWConv} depthwise convolution \textit{FSConv} fully separable convolution, $k$ the filter size of the fully separable convolution, \textit{BN} batch normalization, \textit{ReLU} relu non-linear activation function.) 
}
\label{fig:aggregation-module}
\end{figure}

\subsection{Context Prior Layer}
Context Prior Layer considers an input feature $\mathbold{X}$ with the shape of $H \times W \times C_0$, as illustrated in Figure~\ref{fig:network-structure}.
We adopt an aggregation module to adapt $\mathbold{X}$ to $\mathbold{\widetilde{X}}$ with the shape of $H \times W \times C_1$.
Given $\mathbold{\widetilde{X}}$, one $1 \times 1$ convolution layer followed by a BN layer~\cite{Ioffe-ICML-BN-2015} and a Sigmoid function is applied to learn a prior map $\mathbold{P}$ with the size $H \times W \times N (N=H \times W)$.
With the explicit supervision of the Affinity Loss, Context Prior Map $\mathbold{P}$ can encode the relationship between intra-class pixels and inter-class pixels.
The intra-class is given by $\mathbold{Y} = \mathbold{P} \mathbold{\widetilde{X}}$, where $\mathbold{\widetilde{X}}$ is reshaped into $N \times C_1$ size.
In this operator, the prior map can adaptively select the intra-class pixels as the intra-class context for each pixel in the feature map.
On the other hand, the reversed prior map is applied to selectively highlight the inter-class pixels as the inter-class context: $\mathbold{\overline{Y}} = (\mathds{1} - \mathbold{P}) \mathbold{\widetilde{X}}$, where $\mathds{1}$ is an all-ones matrix with the same size of $\mathbold{P}$.
Finally, we concatenate the original feature and both kinds of context to output the final prediction: $\mathcal{F} = Concat(\mathbold{X}, \mathbold{Y}, \mathbold{\overline{Y}})$.
With both context, we can reason the semantic correlation and scene structure for each pixel.

\subsection{Aggregation Module}
As discussed in Section~\ref{sec:intro}, the Context Prior Map requires some local spatial information to reason the semantic correlation.
Therefore, we devise an efficient Aggregation Module with the fully separable convolution (separate on both the spatial and depth dimensions) to aggregate the spatial information.
The convolution layer can inherently aggregate nearby spatial information.
A natural method to aggregate more spatial information is to use the a large filter size convolutions.
However, convolutions with large filter size are computationally expensive.
Therefore, similar to~\cite{Szegedy-CVPR-Inception-2016, Peng-CVPR-Largekernl-2017}, we factorize the standard convolution into two asymmetric convolutions spatially.
For a $k \times k$ convolution, we can use a $k \times 1$ convolution followed by a $1 \times k$ convolution as the alternative, termed spatial separable convolution.
It can decrease $\frac{k}{2}$ computation and keep the equal size of receptive filed in comparison to the standard convolution. 
Meanwhile, each spatial separable convolution adopts the depth-wise convolution~\cite{Chollet-CVPR-Xception-2017, Zhang-CVPR-Shufflenet-2018, Howard-Arxiv-MobileNet-2017}, further leading to the computation decrease.
We call this separable convolution as Fully Separable Convolution with consideration both the spatial and depth dimensions.
Figure~\ref{fig:aggregation-module} demonstrates the complete structure of the Aggregation Module.

\subsection{Network Architecture}
The Context Prior Network (CPNet) is a fully convolutional network composed of a backbone network and a Context Prior Layer, as shown in Figure~\ref{fig:network-structure}.
The backbone network is an off-the-shelf convolutional network~\cite{He-CVPR-ResNet-2016, Zhang-CVPR-Shufflenet-2018, Wang-TPAMI-HRNet-2019}, \eg ResNet~\cite{He-CVPR-ResNet-2016}, with the dilation strategy~\cite{Zhao-CVPR-PSPNet-2017, Zhao-ECCV-PSANet-2018, Zhang-CVPR-EncNet-2018}.
In the Context Prior Layer, the Aggregation Module first aggregates some spatial information efficiently.
Based on the aggregated spatial information, the Context Prior Layer learns a context prior map to capture intra-class context and inter-class context.
Meanwhile, the Affinity Loss regularizes the learning of Context Prior, while the cross-entropy loss function is  the segmentation supervision.
Following the pioneering work~\cite{Zhao-CVPR-PSPNet-2017, Zhao-ECCV-PSANet-2018, Zhang-CVPR-EncNet-2018}, we employ the auxiliary loss on stage $4$ of the backbone network, which is also a cross-entropy loss.
The final loss function is as follows:
\begin{equation}
	\mathcal{L} = \lambda_s \mathcal{L}_{s} + \lambda_a \mathcal{L}_{a} + \lambda_p \mathcal{L}_{p},	
\end{equation}
where $\mathcal{L}_{s}$, $\mathcal{L}_{a}$, and $\mathcal{L}_{p}$ represent the main segmentation loss, auxiliary loss, and affinity loss functions, respectively.
In addition, $\lambda_s$, $\lambda_a$, and $\lambda_p$ are the weights to balance the segmentation loss, auxiliary loss, and affinity loss, respectively.
We empirically set the weights as: $\lambda_s=1$ and $\lambda_p=1$.
Similar to \cite{Zhao-CVPR-PSPNet-2017, Zhao-ECCV-PSANet-2018, Zhang-CVPR-EncNet-2018}, we set the weight: $\lambda_a=0.4$,

\section{Experimental Results}
In this section, we first introduce the implementation and training details of the proposed network.
Next, we evaluate the proposed method and compare it with \emph{state-of-the-art} approaches on three challenging scene segmentation datasets, including ADE20K~\cite{Zhou-ADE-2017}, PASCAL-Context~\cite{PASCAL-Context}, and Cityscapes~\cite{Cityscapes}.
We implement the proposed model using PyTorch~\cite{PyTorch} toolbox.

\subsection{Implementation Details}
\paragraph{Network.}
We adopt the ResNet~\cite{He-CVPR-ResNet-2016} as our pre-trained model with dilation strategy~\cite{Chen-ICLR-Deeplabv2-2016, Chen-Arxiv-Deeplabv3-2017, Chen-ECCV-Deeplabv3p-2018}.
Then we adopt the bilinear interpolation to up-sample the prediction eight times to compute the segmentation loss.
Following~\cite{Zhao-CVPR-PSPNet-2017, Zhao-ECCV-PSANet-2018, Zhang-CVPR-EncNet-2018}, we integrate the auxiliary loss on stage 4 of the backbone network.
We set the filter size of the fully separable convolution in the Aggregation Module as 11.
\paragraph{Data Augmentation.}
In the training phase, we apply the mean subtraction, random horizontal flip and random scale, which contains \{0.5, 0.75, 1.0, 1.5, 1.75, 2.0\}, on the input images in avoiding of overfitting.
Finally, we randomly crop the large image or pad the small image into a fix size for training ($480\times480$ for ADE20K, $512\times512$ for PASCAL-Context and $768\times768$ for Cityscapes).
\paragraph{Optimization.}
We fine-tune the CPNet model using the stochastic gradient descent (SGD) algorithm~\cite{Krizhevsky-NIPS-Imagenet} with 0.9 momentum, $10^{-4}$ weight decay and 16 batch size. 
Notably, we set the weight decay as $5\times10^{-4}$ when training on the Cityscapes dataset.
Following the pioneering work~\cite{Chen-Arxiv-Deeplabv2-2016, Chen-Arxiv-Deeplabv3-2017, Yu-CVPR-DFN-2018, Yu-ECCV-BiSeNet-2018}, we adopt the ``poly'' learning rate strategy $\gamma = {{\gamma}_0} \times (1 - \frac{N_{iter}}{N_{total}})^{p}$, where $N_{iter}$ and $N_{total}$ represent the current iteration number and total iteration number, and $p=0.9$.
We set the base learning rate ${\gamma}_0$ as $2\times10^{-2}$ for the experiments on ADE20K, while $1\times10^{-2}$ for the experiments on PASCAL-Context and Cityscapes.
Meanwhile, we train the model for 80K iterations on ADE20K, 25K for PASCAL-Context and 60K for Cityscapes.
We use the standard cross entropy loss when training on the ADE20K and PASCAL-Context dataset.
While training on Cityscapes, similar to \cite{Wu-Arxiv-HighPerf-2016, Yu-ECCV-BiSeNet-2018, Yuan-Arxiv-OCNet-2018}, we adopt the bootstrapped cross-entropy loss~\cite{Wu-Arxiv-HighPerf-2016} to mitigate the class imbalance problem in this dataset.
\paragraph{Inference.}
In the inference phase, following~\cite{Zhao-CVPR-PSPNet-2017, Peng-CVPR-Largekernl-2017, Yu-CVPR-DFN-2018, Zhang-CVPR-EncNet-2018}, we average the predictions of multiple scaled and flipped inputs to further improve the performance.
We use the scales including \{0.5, 0.75, 1.0, 1.5, 1.75\} for the ADE20K and PASCAL-Context datasets, while \{0.5, 0.75, 1, 1.5\} for the Cityscapes dataset.
In addition, we adopt the pixel accuracy (pixAcc) and mean intersection of union (mIoU) as the evaluation metrics.

\subsection{Evaluations on the ADE20K Dataset}
\paragraph{Dataset description.} 
ADE20K is a challenging scene parsing benchmark due to its complex scene and up to 150 category labels. 
This dataset can be divided into 20K/2K/3K for training, validation and testing respectively.
We report the results on the validation set using pixAcc and mIoU.

\begin{table}[t]
\centering
\small
\tablestyle{6pt}{1.05}
\begin{tabular}{l|x{22}x{22}}
\shline
\multicolumn{1}{c|}{model} & \emph{mIoU} & \emph{pixAcc} \\
\hline
ResNet-50 (Dilation) &  				34.38 & 76.51 \\
ResNet-50 + Aux (Baseline) &  36.24 & 77.37 \\
ResNet-50 + ASPP &  		40.39 & 79.71 \\
ResNet-50 + PSP &  			41.49 & 79.61 \\
ResNet-50 + NonLocal &  	40.96 & 79.98 \\
ResNet-50 + PSA &			41.92 & 80.17 \\
\hline
ResNet-50 + Aggregation Module & 41.51 & 79.93 \\
ResNet-50 + IntraPrior (BCE) &  42.34 & 80.15  \\
ResNet-50 + InterPrior (BCE) &  41.88 & 79.96 \\
ResNet-50 + IntraPrior (AL) &  	42.74 & 80.30 \\
ResNet-50 + InterPrior (AL) &  	42.43 & 80.21 \\
ResNet-50 + ContextPriorLayer &  	43.92 & 80.77 \\
ResNet-50 + ContextPriorLayer\_MS &  44.46 & \underline{81.38}  \\
ResNet-101 + ContextPriorLayer &     \underline{45.39} & 81.04 \\
ResNet-101 + ContextPriorLayer\_MS & \textbf{46.27} & \textbf{81.85} \\
\shline
\end{tabular}
\vspace{0.5em}
\caption{Ablative studies on the ADE20K~\cite{Zhou-ADE-2017} validation set in comparison to other contextual information aggregation approaches.
(Notation:
\textit{Aux} auxiliary loss, \textit{BCE} binary cross entropy loss, \textit{AL} Affinity Loss, \textit{MS} multi-scale and flip testing strategy.)
}
\label{tab:res-ade}
\end{table}

\paragraph{Ablation studies.}
To demonstrate the effectiveness of our Context Prior and CPNet, we conduct the experiments with different settings and compared with other spatial information aggregation module, as shown in Table~\ref{tab:res-ade}.

First, we introduce our baseline model.
We evaluate the FCN~\cite{Long-CVPR-FCN-2015} model with dilated convolution~\cite{Chen-ICLR-Deeplabv2-2016} based on ResNet-50~\cite{He-CVPR-ResNet-2016} on the validation set.
Following \cite{Zhao-CVPR-PSPNet-2017, Zhang-CVPR-EncNet-2018, Zhao-ECCV-PSANet-2018}, we add the auxiliary loss on stage 4 of the ResNet backbone.
This can improve mIoU by $1.86\%$ ($34.38\%\rightarrow36.24\%$) and pixAcc by  $0.86\%$ ($76.51\%\rightarrow77.37\%$).
We adopt this model as our baseline.

Based on the features extracted by FCN, various methods aggregate contextual information to improve the performance.
The pyramid-based methods (\eg PSP and ASPP) adopts pyramid pooling or pyramid dilation rates to aggregate multi-range spatial information.
Recent approaches~\cite{Yuan-Arxiv-OCNet-2018, Fu-CVPR-DANet-2019} apply the self-attention~\cite{Wang-CVPR-Nonlocal-2018} method to aggregate the long-range spatial information, while the PSA module~\cite{Zhao-ECCV-PSANet-2018} learns over-parametric point-wise attention.
Table~\ref{tab:res-ade} lists our reimplement results with different spatial information aggregation modules.
While these methods can improve the performance over the baseline, they aggregate the spatial information as a mixture of the intra-class and inter-class context, maybe making the network confused, as discussed in Section~\ref{sec:intro}.
Therefore, different from these methods, the proposed CPNet considers the contextual dependencies as a Context Prior to encoding the identified contextual relationship.
Specifically, for each pixel, we capture the intra-class context and inter-class context with the Context Prior Layer.
With the same backbone ResNet-50 and without other testing tricks, our method performs favorably against these methods. 

We also investigate the effectiveness of the Aggregation Module, IntraPrior branch, InterPrior branch and Affinity Loss in our CPNet model.
We use the Aggregation Module with filter size $11$ to aggregate the local spatial information.
Similar to \cite{Zhao-ECCV-PSANet-2018}, the Aggregation Module generates an attention mask with the resolution of $N \times N (N = H \times W)$ to refine the prediction.
As shown in Table~\ref{tab:res-ade}, the Aggregation Module improves the mIoU and pixAcc by $5.27\%$ / $2.56\%$ over the baseline model.
With the IntraPrior branch based on the binary cross entropy loss, our single scale testing results obtain $42.34\%$ / $80.15\%$ in terms of mIoU and pixAcc, surpassing the baseline by $6.1\%$ / $2.78\%$.
On the other hand, the InterPrior branch achieves $42.88\%$ / $79.96\%$ with the same setting.
Both of the significant improvements demonstrate the effectiveness of the proposed Context Prior.

\begin{figure}[t]
\footnotesize
\centering
\renewcommand{\tabcolsep}{1pt} %
\ra{1} %
\begin{center}
\begin{tabular}{cccc}
\includegraphics[width=0.24\linewidth]{ade-res/img1.jpg} &
\includegraphics[width=0.24\linewidth]{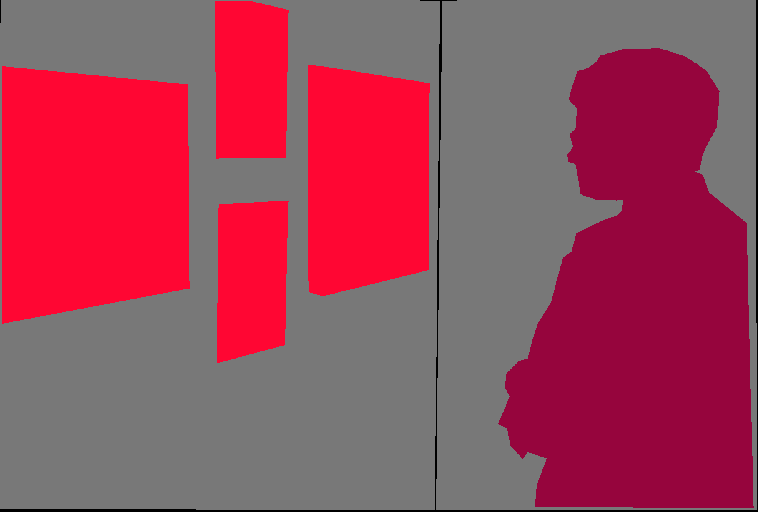} &
\includegraphics[width=0.24\linewidth]{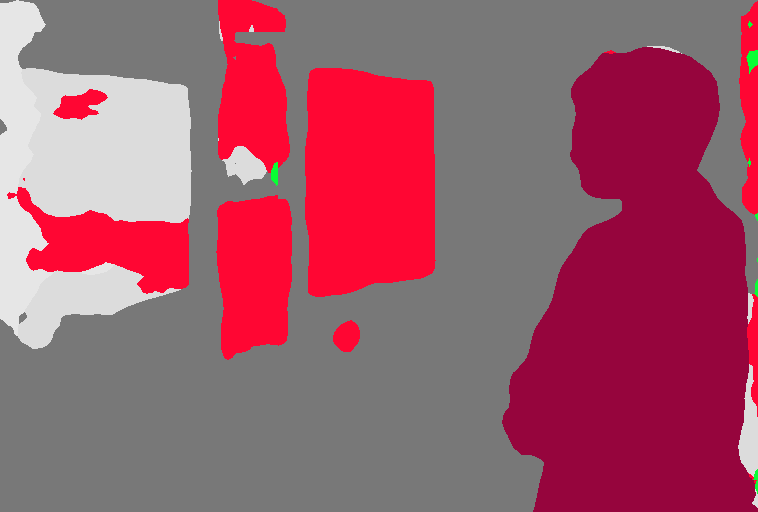} &
\includegraphics[width=0.24\linewidth]{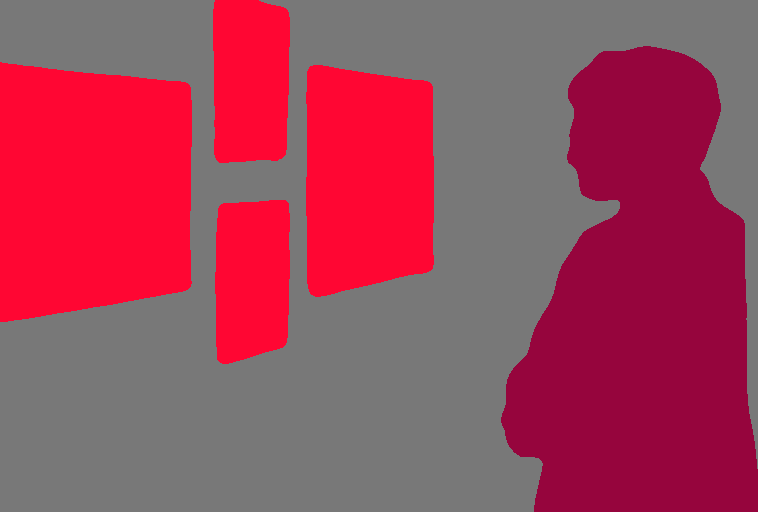} \\
\includegraphics[width=0.24\linewidth]{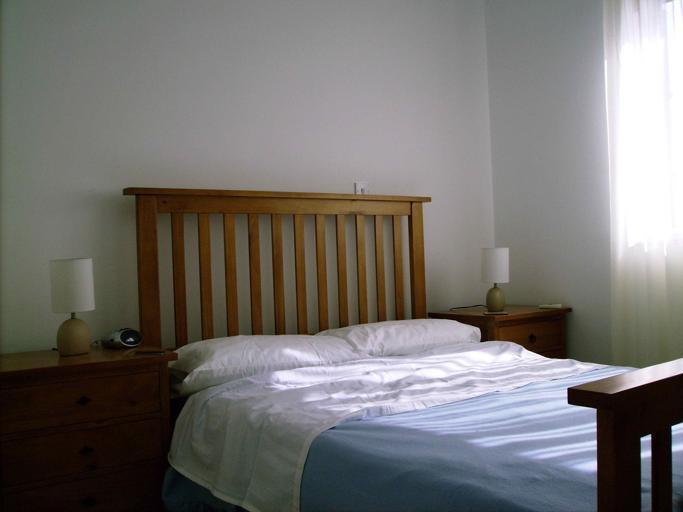} &
\includegraphics[width=0.24\linewidth]{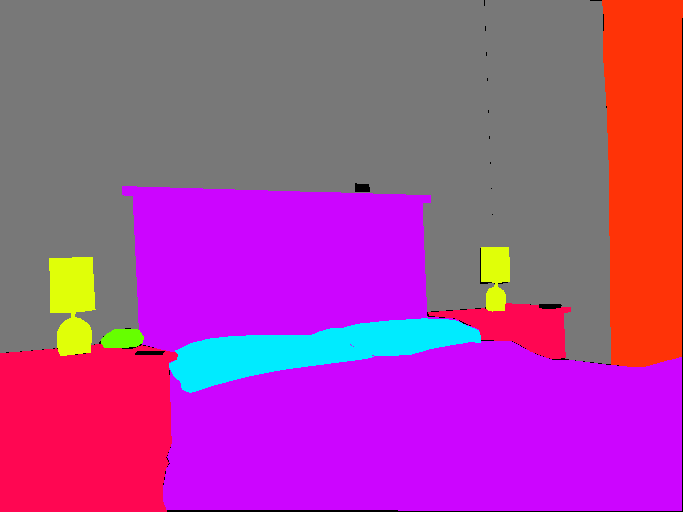} &
\includegraphics[width=0.24\linewidth]{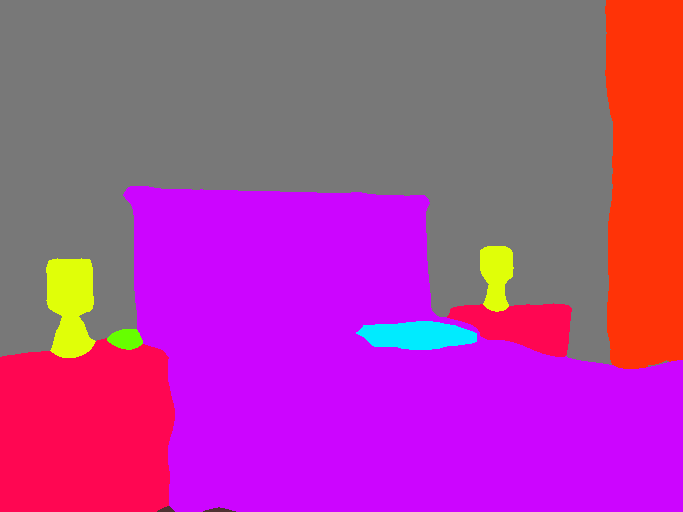} &
\includegraphics[width=0.24\linewidth]{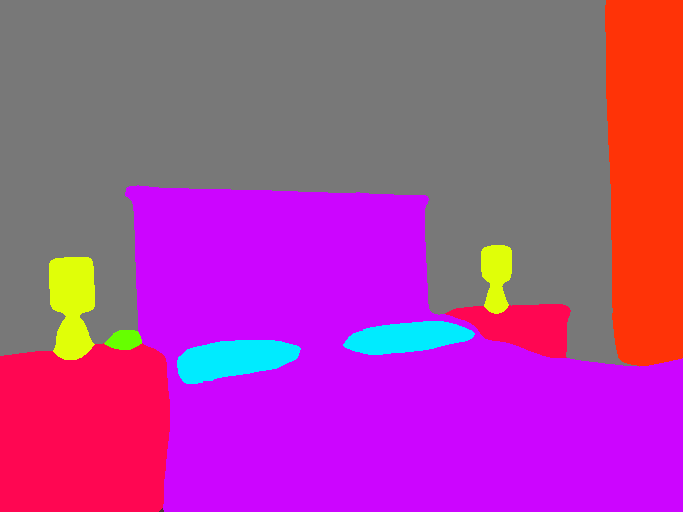} \\
\includegraphics[width=0.24\linewidth]{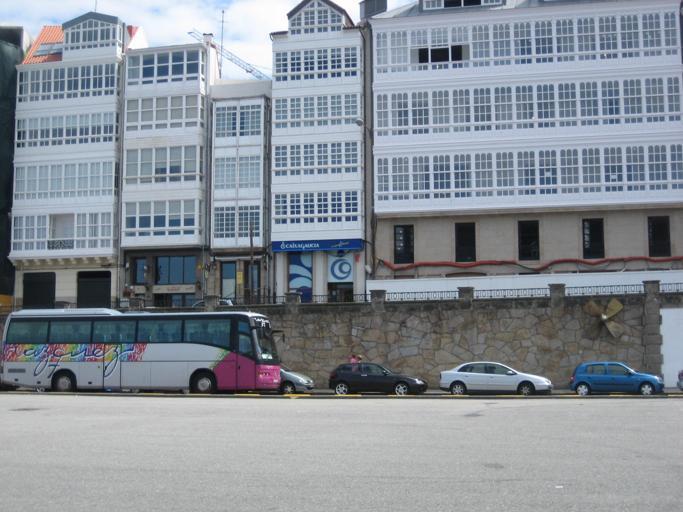} &
\includegraphics[width=0.24\linewidth]{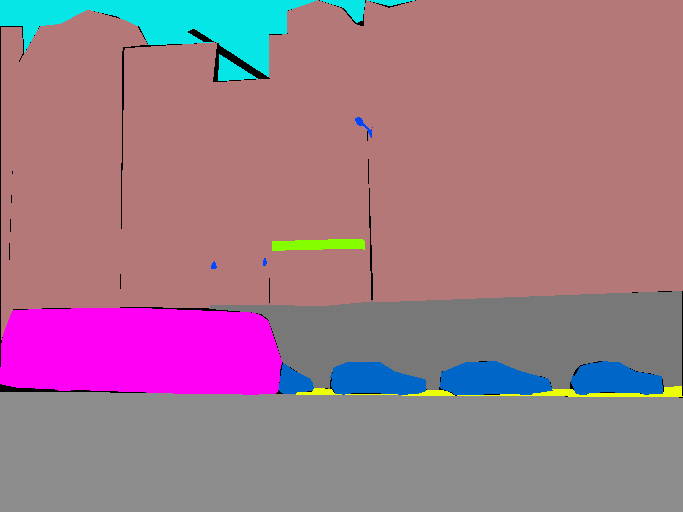} &
\includegraphics[width=0.24\linewidth]{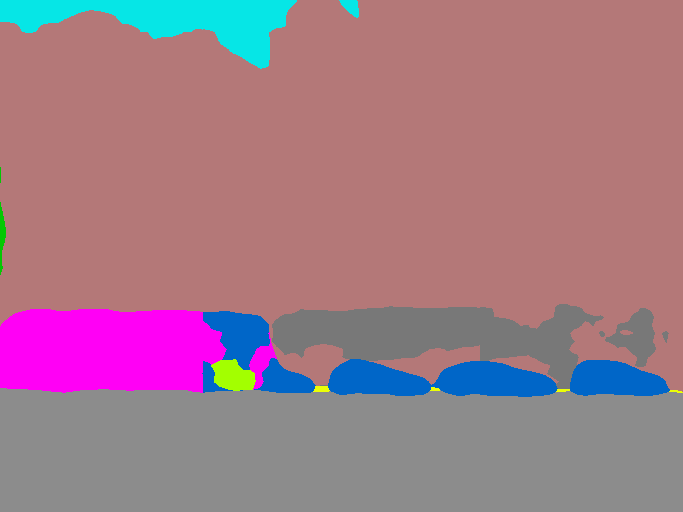} &
\includegraphics[width=0.24\linewidth]{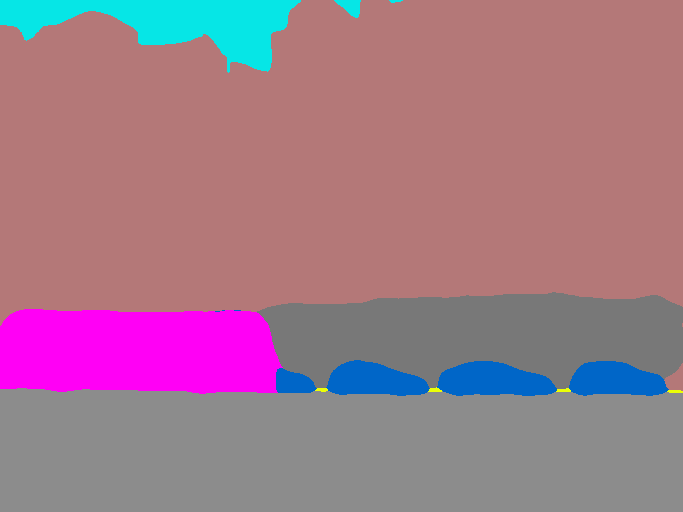} \\
\includegraphics[width=0.24\linewidth]{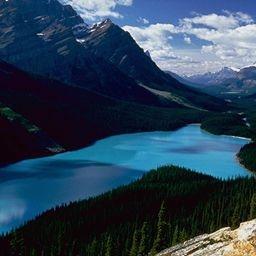} &
\includegraphics[width=0.24\linewidth]{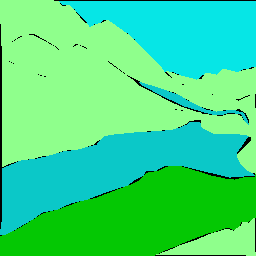} &
\includegraphics[width=0.24\linewidth]{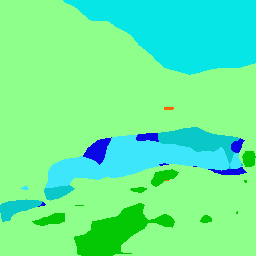} &
\includegraphics[width=0.24\linewidth]{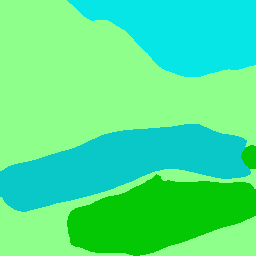} \\
(a) Input Image &
(b) Ground Truth &
(c) FCN &
(d) CPNet (ours) \\
\end{tabular}
\end{center}
\caption{
\textbf{Visual improvement} on validation set of ADE20K.
Harvesting the intra-class context and inter-class context is helpful to the scene understanding.
}
\label{fig:ade-vis}
\end{figure}

To further improve the quality of the context prior map, we devise an Affinity Loss.
Table~\ref{tab:res-ade} indicates that the Affinity Loss can improve the mIoU and pixAcc by $0.4\%$ / $0.15\%$ based on IntraPrior branch, while boosting $0.55\%$ / $0.25\%$ based on InterPrior branch.
We integrate both IntraPrior branch and InterPrior branch with the Affinity Loss to achieve $43.92\%$ mIoU and $80.77\%$ pixAcc, which demonstrates that both priors can be complementary.
To further improve the performance, we apply the multi-scale and flipped testing strategy to achieve $44.46\%$ mIoU and $81.38\%$ pixAcc.
Deeper network leading to better feature representation, our CPNet obtains $45.39\%$ mIoU and $81.04\%$ pixAcc with the ResNet-101.
With the testing strategy, our model based on ResNet-101 achieves $46.27\%$ mIoU and $81.85\%$ pixAcc.
Figure~\ref{fig:ade-vis} provides some visualization examples.

\begin{table}[t]
\centering
\small
\tablestyle{6pt}{1.05}
\begin{tabular}{l|x{15}x{15}x{15}x{15}x{15}x{15}x{15}}
\shline
\multicolumn{1}{c|}{$k$}     & 3     & 5     & 7     & 9     & 11    & 13 & 15 \\
\hline
w/o CP & 42.06 & 41.86 & 41.87 & 42.32 & 41.51 & \textbf{42.34}   & 42.23    \\
w/ CP   & 42.26 & 42.81 & 43.38 & 43.14 & \textbf{43.92} & 42.54   & 42.59   \\
\hline
\multicolumn{1}{c|}{$\Delta$}  & 0.2   & 0.95  & 1.51  & 0.82  & \textbf{2.41}  & 0.2   & 0.36   \\
\shline
\end{tabular}
\vspace{0.5em}
\caption{Experimental results (mIoU) w/ or w/o Context Prior based on different kernel sizes.
(Notation:
\textit{$k$} the kernel size of the fully separable convolution, \textit{$\Delta$} the improvement of introducing the Context Prior, \textit{CP} Context Prior.)
}
\label{tab:comp-ks}
\end{table}

\begin{table}[t]
\centering
\small
\tablestyle{6pt}{1.05}
\begin{tabular}{l|x{22}x{22}x{22}}
\shline
 & PPM & ASPP & AM\\
\hline
w/o CP & 41.49 & 40.39 & \textbf{41.51}\\
w/ CP & 42.55 & 42.69  & \textbf{43.92}\\
\hline
\multicolumn{1}{c|}{$\Delta$} & $\uparrow$1.06 & $\uparrow$2.3 & $\uparrow$\textbf{2.41}\\
\shline
\end{tabular}
\caption{Generalization to the PPM and ASPP module.
The evaluation metric is mIoU (\%).
(Notation: \textit{PPM} pyramid pooling module, \textit{ASPP} atrous spatial pyramid pooling, \textit{CP} Context Prior, \textit{AM}: Aggregation Module.)
}
\vspace{0.5em}
\label{tab:comp-cp-psp-aspp}
\end{table}

\paragraph{Analysis and discussion.}
In Table~\ref{tab:res-ade}, the proposed CPNet achieves 
considerable 
improvement on the ADE20K benchmark.
Someone may argue that the large filter size of the Aggregation Module leads to the performance gain.
Or one may question whether the Context Prior can generalize to other algorithms.
We thus
provide more evidence to thoroughly understand the Context Prior.
We conduct the discussion experiments on the ADE20K validation set with ResNet-50 backbone.
The results reported in Table~\ref{tab:comp-ks} and Table~\ref{tab:comp-cp-psp-aspp} are the single scale testing results.

(1) \textit{The influence between the spatial information and Context Prior.}
As discussed in Section~\ref{sec:context-prior}, the distinguished contextual dependencies are helpful to scene understanding.
Therefore, we propose a Context Prior to model the intra-context and inter-context.
Meanwhile, the Context Prior requires some spatial information to reason the relationship.
To this end, we integrate an Aggregation Module in the Context Prior Layer.

Table~\ref{tab:comp-ks} indicates that with the increasing filter size, the models without Context Prior obtain the close results.
However, with Context Prior, each model achieves improvements steadily. 
Meanwhile, the improvements gradually increase with the increasing filter size.
When the filter size is 11, the performance ($43.92\%$ mIoU) and the relative gain ($2.41\%$) reach the peak.
If we continue to increase the filter size, the performance and the corresponding improvement both drop.
In other words, Context Prior requries appropriate local spatial information to reason the relationships.

(2) \textit{Generalization to other spatial information aggregation module.}
To validate the generalization ability of the proposed Context Prior, we further replace the Aggregation Module with PPM or ASPP module to generate Context Prior Map with the supervision of Affinity Loss.
As shown in Table~\ref{tab:comp-cp-psp-aspp}, Context Prior can further improve the mIoU by $1.06\%$ over the PPM without Context Prior,  $2.3\%$ over the ASPP module and $2.41\%$ over our Aggregation Module.
This improvement demonstrates the effectiveness and generalization ability of our Context Prior.
Besides, without Context Prior, our Aggregation Module also achieves the highest performance comparing to the PPM and ASPP module.

\begin{figure}[t]
\footnotesize
\centering
\renewcommand{\tabcolsep}{1pt} %
\ra{1} %
\begin{center}
\begin{tabular}{ccc}
\includegraphics[width=0.33\linewidth]{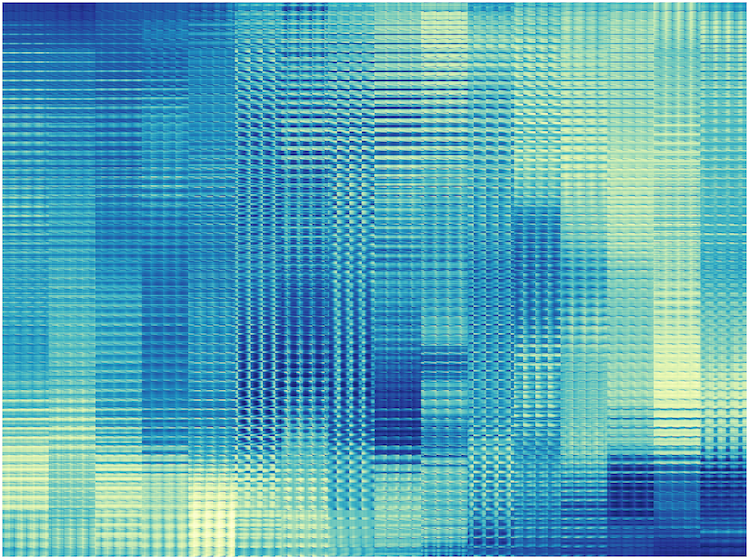} &
\includegraphics[width=0.33\linewidth]{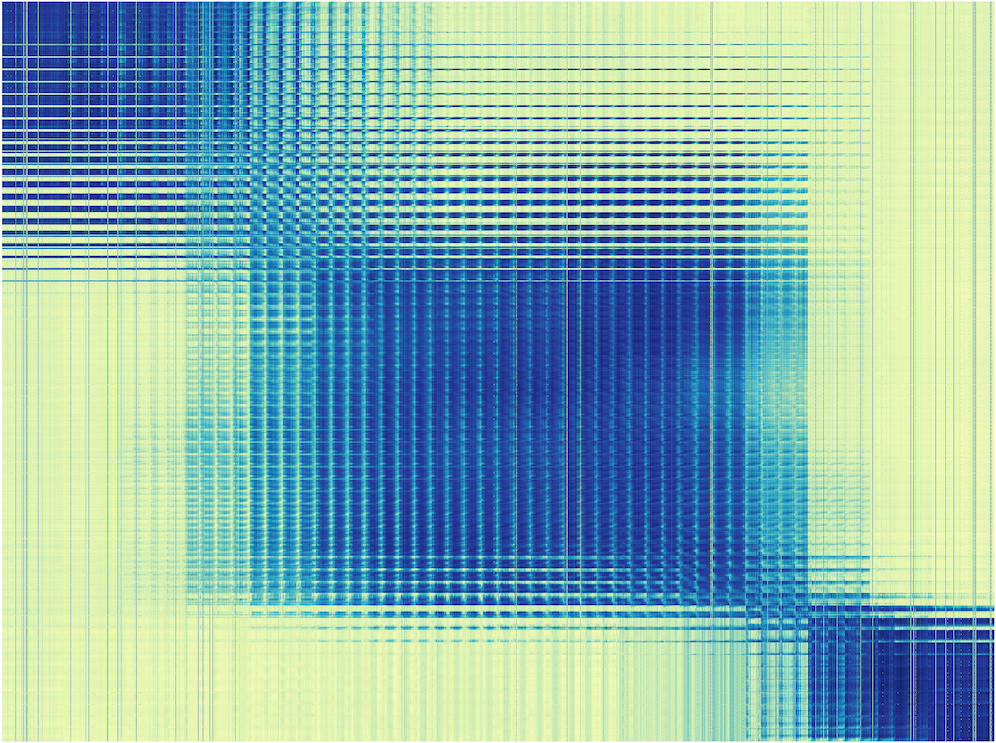} &
\includegraphics[width=0.33\linewidth]{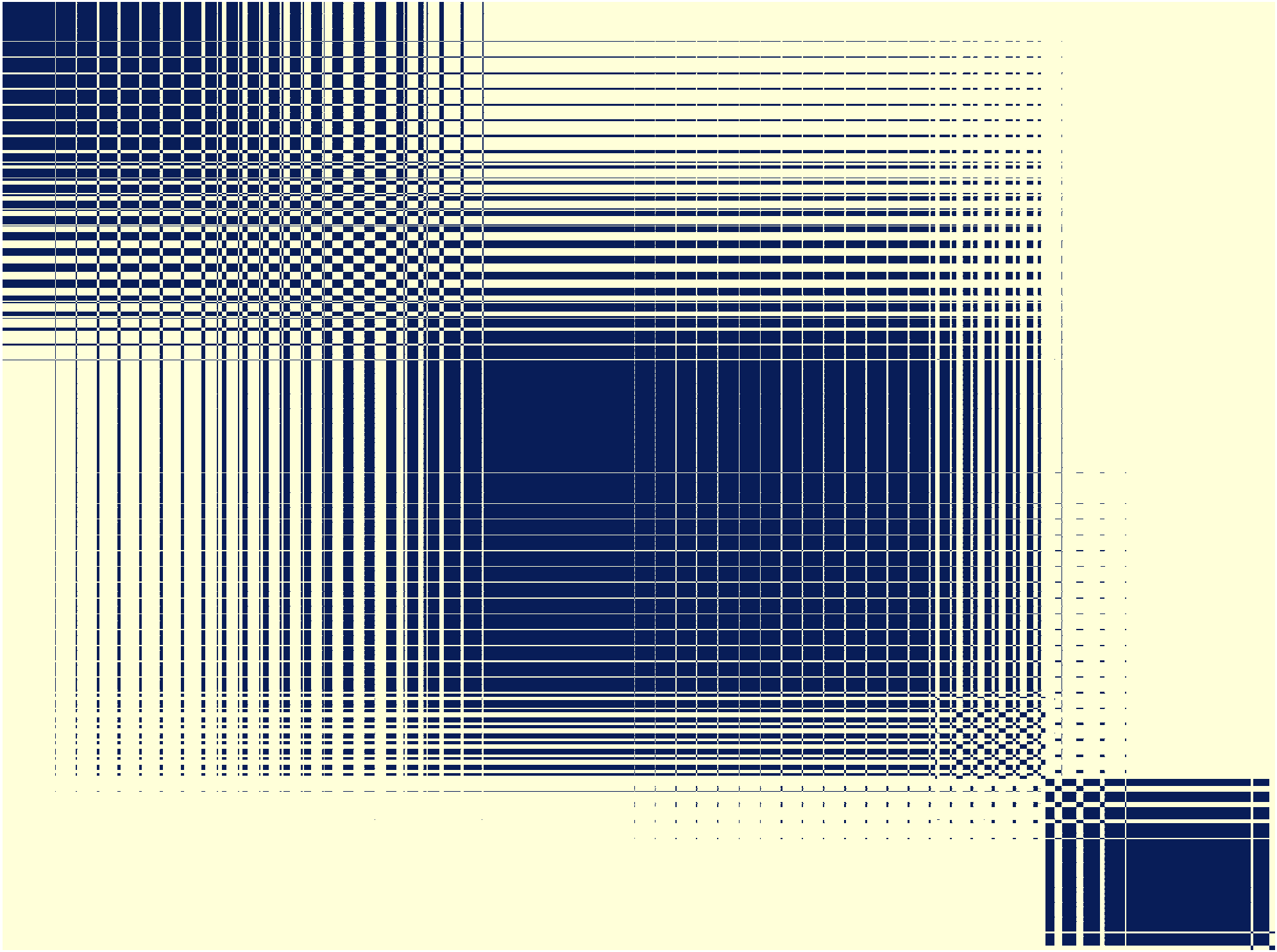} \\
\includegraphics[width=0.33\linewidth]{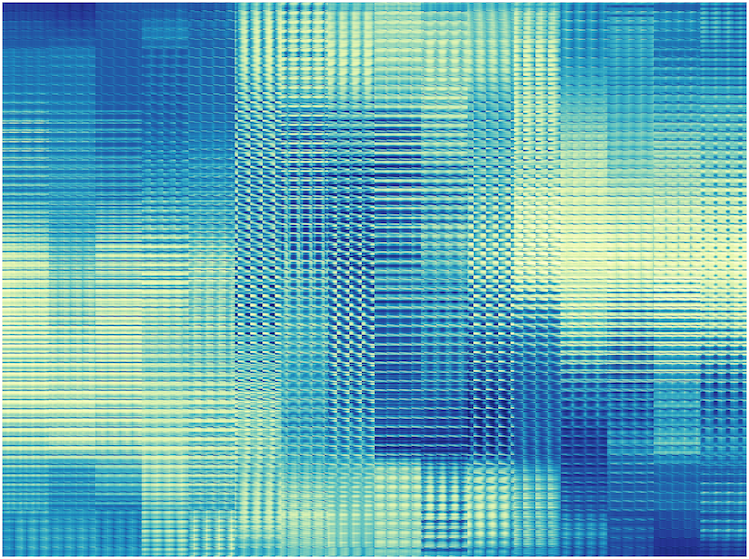} &
\includegraphics[width=0.33\linewidth]{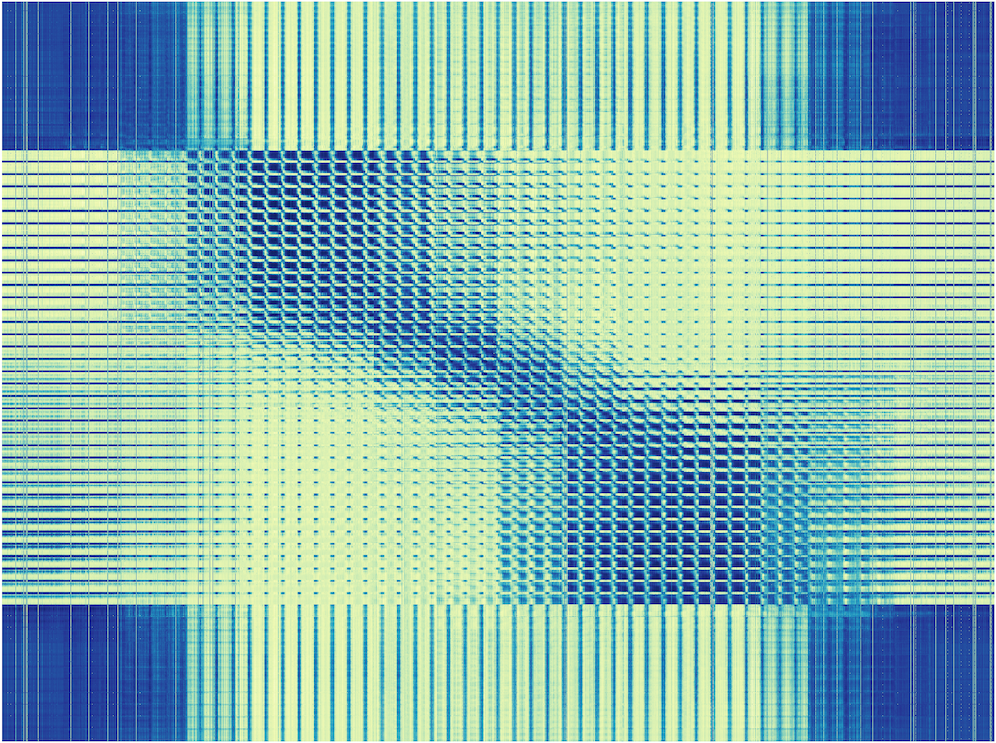} &
\includegraphics[width=0.33\linewidth]{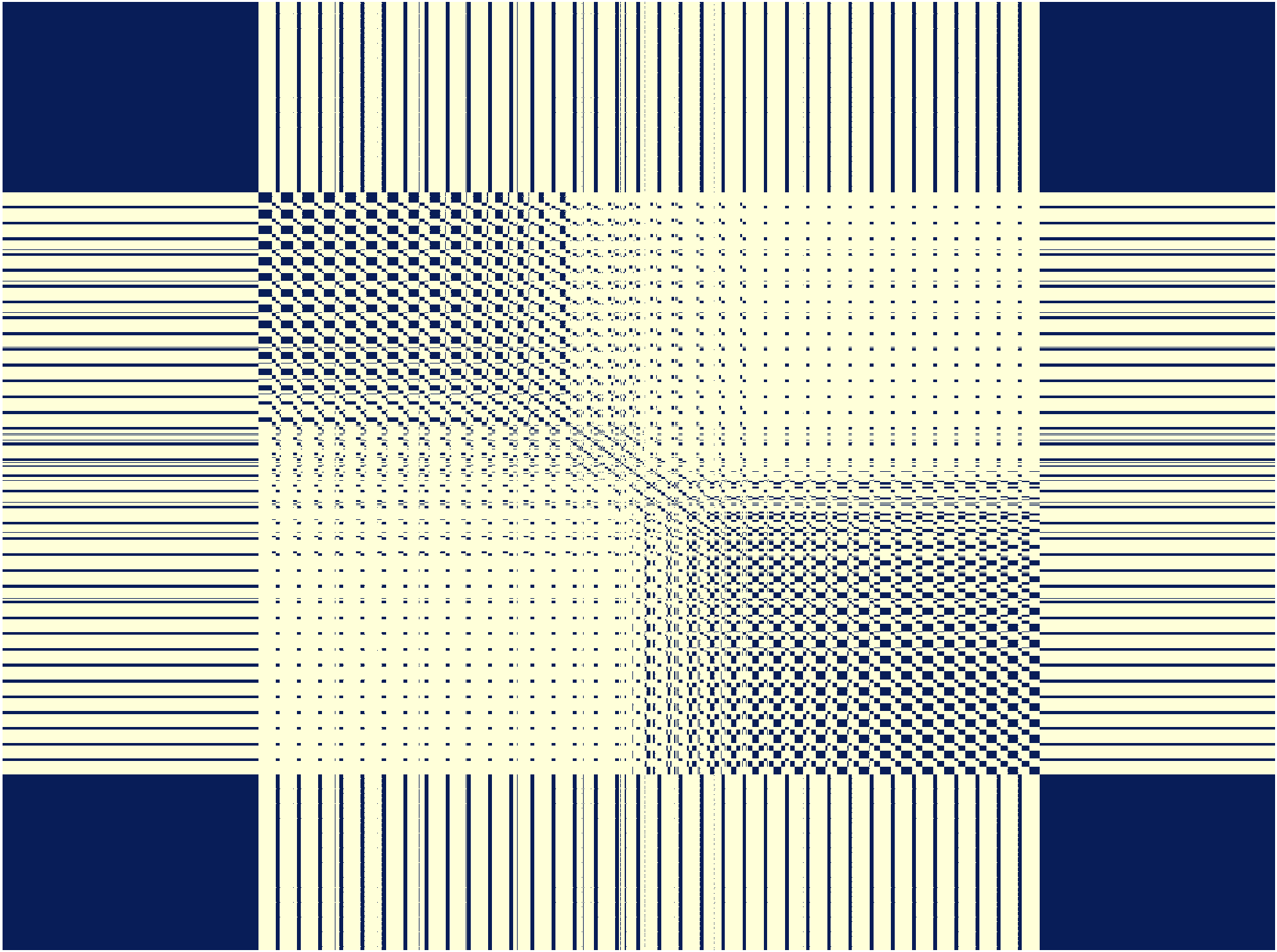} \\
\includegraphics[width=0.33\linewidth]{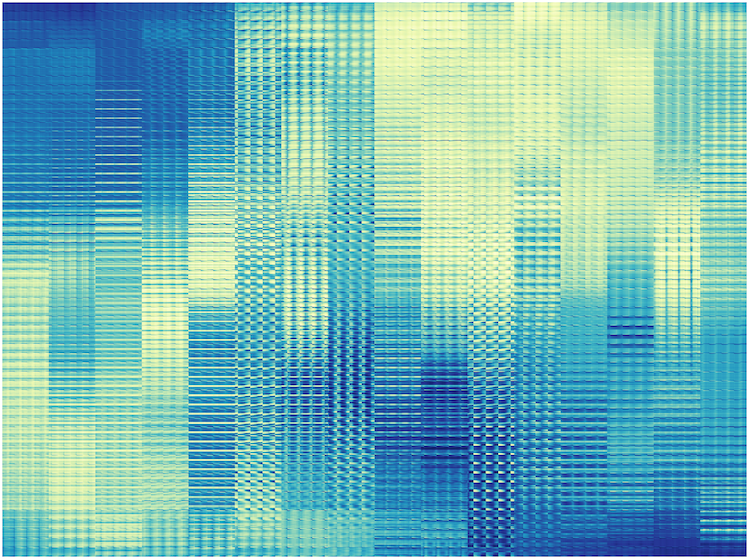} &
\includegraphics[width=0.33\linewidth]{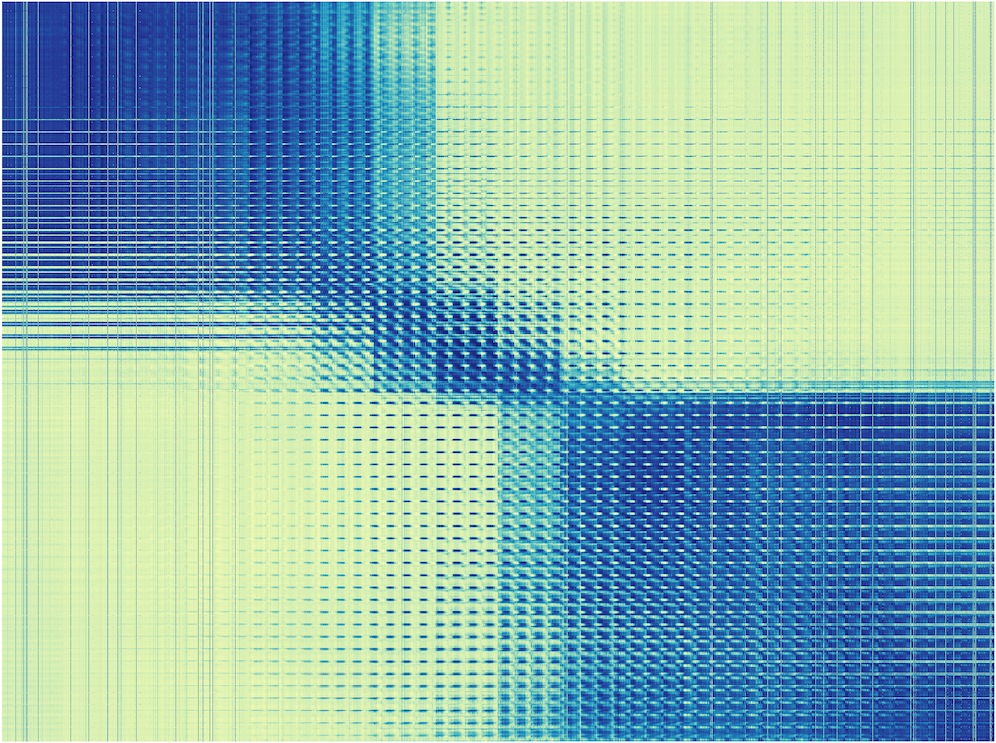} &
\includegraphics[width=0.33\linewidth]{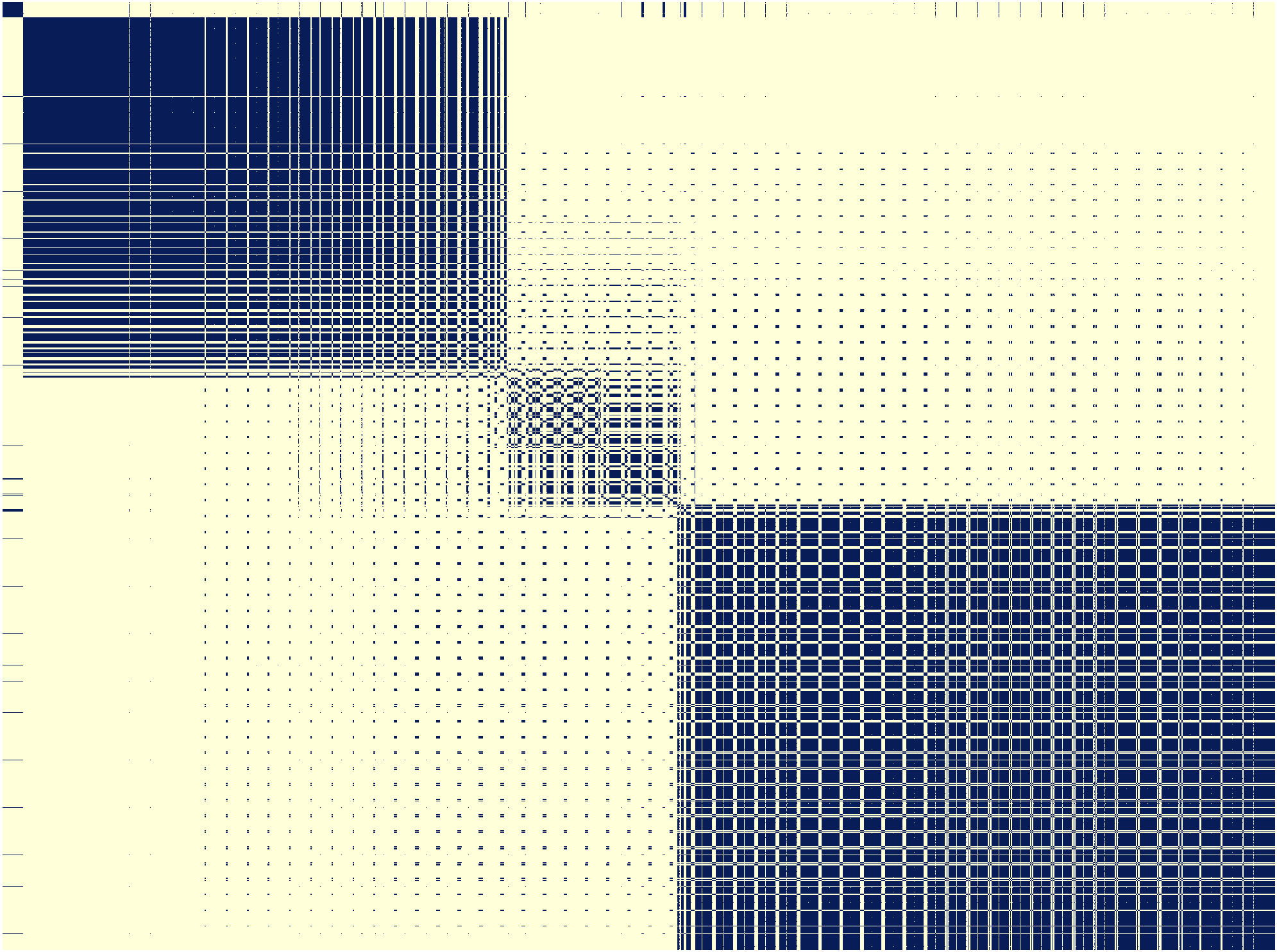} \\
(a) Attention Map &
(b) Learned Prior Map &
(c) Ideal Affinity Map\\
\end{tabular}
\end{center}
\caption{{\bf Visualization of the Prior Map predicted by our CPNet}.
(a) We only 
use
the Aggregation Module to generate an attention map without the supervision of the Affinity Loss.
(b) With the guidance of the Affinity Loss, the Context Prior Layer can capture the intra-class context and inter-class context.
(c) The Ideal Affinity Map is constructed from the ground truth.
Deeper color denotes higher response.
}
\label{fig:prior-map}
\end{figure}

\paragraph{Visualization of prior maps.}
To get a deeper understanding of our Context Prior, we randomly choose some examples from the ADE20K validation set and visualize the learned Context Prior Maps in Figure~\ref{fig:prior-map}.
We use the Aggregation Module to generate the attention map without the guidance of the Affinity Loss.
Compared with the Ideal Affinity Map, we observe this attention map actually has a rough trend to learn this relationship.
With the Affinity Loss, our Context Prior Layer can learn a prior map with more explicit structure information, which helps to refine the prediction.

\paragraph{Comparison with state-of-the-art.}
We conduct the comparison experiments with other \emph{state-of-the-art} algorithms on Table~\ref{tab:comp-ade}.
The proposed CPNet achieves $46.27\%$ mIoU and $81.85\%$ pixAcc, which performs favorably against previous \emph{state-of-the-art} methods, even exceeds the winner entry of the COCO-Place Challenge 2017 based on ResNet-269.
Our CPNet50 (with ResNet-50 as the backbone) achieves $44.46\%$ mIoU and $81.38\%$ pixAcc, even outperforms PSPNet~\cite{Zhao-CVPR-PSPNet-2017}, PSANet~\cite{Zhao-ECCV-PSANet-2018} and SAC~\cite{Zhang-ICCV-SAC-2017} with deeper ResNet-101 and RefineNet with much deeper ResNet-152 as the backbone.
This significant improvement manifests the effectiveness of our Context Prior.

\begin{table}[t]
\centering
\small
\tablestyle{6pt}{1.05}
\begin{tabular}{l|c|l|x{22}|x{22}}
\shline
\multicolumn{1}{c|}{model} & reference & \multicolumn{1}{c|}{backbone} & \emph{mIoU} & \emph{picAcc} \\
\hline
RefineNet~\cite{Lin-CVPR-Refinenet-2017} & CVPR2017 & ResNet-101 & 40.2  &  -     \\
RefineNet~\cite{Lin-CVPR-Refinenet-2017} & CVPR2017 & ResNet-152 & 40.7  &  -     \\
UperNet~\cite{Xiao-UperNet-ECCV-2018} 	 & ECCV2018 & ResNet-101 & 42.66 &  81.01 \\
PSPNet~\cite{Zhao-CVPR-PSPNet-2017}      & CVPR2017 & ResNet-101 & 43.29 &  81.39 \\
PSPNet~\cite{Zhao-CVPR-PSPNet-2017}      & CVPR2017 & ResNet-269 & 44.94 &  81.69 \\
DSSPN~\cite{Liang-DSSPN-CVPR-2018}       & CVPR2018 & ResNet-101 & 43.68 &  81.13 \\
PSANet~\cite{Zhao-ECCV-PSANet-2018}      & ECCV2018 & ResNet-101 & 43.77 &  81.51 \\
SAC~\cite{Zhang-ICCV-SAC-2017} 			 & ICCV2017 & ResNet-101 & 44.30 &  \textbf{81.86} \\
EncNet~\cite{Zhang-CVPR-EncNet-2018}     & CVPR2018 & ResNet-101 & 44.65 &  81.69 \\
CFNet~\cite{Zhang-CVPR-ACF-2019} 		 & CVPR2019	& ResNet-101 & 44.89 & - \\
ANL~\cite{Zhu-ICCV-ANL-2019}  			 & ICCV2019	& ResNet-101 & \underline{45.24} & - \\
\hline
CPNet50  & - & ResNet-50   & 44.46 & 81.38 \\
CPNet101 & - & ResNet-101  & \textbf{46.27} & \underline{81.85} \\
\shline
\end{tabular}
\vspace{0.5em}
\caption{Quantitative evaluations on the ADE20K validation set. 
The proposed CPNet performs favorably against \emph{state-of-the-art} segmentation algorithms.}
\label{tab:comp-ade}
\end{table}

\begin{table}[t]
\centering
\small
\tablestyle{6pt}{1.05}
\begin{tabular}{l|c|l|x{22}}
\shline
\multicolumn{1}{c|}{model}  & reference & \multicolumn{1}{c|}{backbone} & \emph{mIoU} \\
\hline
FCN-8S~\cite{Long-CVPR-FCN-2015} 	 				   & CVPR2015 & VGG16 & 37.8 \\
CRF-RNN~\cite{Zheng-ICCV-CRFasRNN-2015} 			   & ICCV2015 & VGG16 & 39.3 \\
BoxSup~\cite{Dai-ICCV-BoxSup-2015}  				   & ICCV2015 & VGG16 & 40.5 \\
Deeplabv2$^{\dagger}$~\cite{Chen-ICLR-Deeplabv2-2016}  & ICLR2016 & ResNet101  & 45.7 \\
RefineNet~\cite{Lin-CVPR-Refinenet-2017}         	   & CVPR2017 & ResNet-152 & 47.3 \\
PSPNet~\cite{Zhao-CVPR-PSPNet-2017} 		           & CVPR2017 & ResNet-101 & 47.8 \\
CCL~\cite{Ding-CVPR-CCL-2018} 					       & CVPR2018 & ResNet-101 & 51.6 \\
EncNet~\cite{Zhang-CVPR-EncNet-2018} 				   & CVPR2018 & ResNet-101 & 51.7 \\
DANet~\cite{Fu-CVPR-DANet-2019} 					   & CVPR2019 & ResNet-101 & 52.6 \\
ANL~\cite{Zhu-ICCV-ANL-2019} 						   & ICCV2019 & ResNet-101 & \underline{52.8} \\
\hline
CPNet101 & - & ResNet-101 	   & \textbf{53.9}     \\
\shline
\end{tabular}
\vspace{0.5em}

\caption{
Quantitative evaluations on the PASCAL-Context validation set. 
The proposed CPNet performs favorably against \emph{state-of-the-art} segmentation methods.
$\dagger$ means the method uses extra dataset.
}
\label{tab:res-context}
\end{table}

\subsection{Evaluations on PASCAL-Context}
\paragraph{Dataset description.}
PASCAL-Context~\cite{PASCAL-Context} is a scene understanding dataset which contains $10,103$ images from PASCAL VOC 2010.
These images are re-annotated as pixel-wise segmentation maps with consideration of both the stuff and thing categories.
This dataset can be divided into $4,998$ images for training and $5,105$ images for testing.
The most common $59$ categories are used for evaluation.

\paragraph{Comparison with state-of-the-art.}
Table~\ref{tab:res-context} shows the performance comparison with other \emph{state-of-the-art} approaches.
Our algorithm achieves $53.9\%$ mIoU on validation set and outperforms \emph{state-of-the-art} EncNet by over $1.0$ point.
Similar to \cite{Chen-ICLR-Deeplabv2-2016, Lin-CVPR-Refinenet-2017, Zhao-CVPR-PSPNet-2017, Ding-CVPR-CCL-2018, Zhang-CVPR-EncNet-2018, Fu-CVPR-DANet-2019}, we evaluate the model with the multi-scale and flipped testing strategy.
The scales contain \{0.5, 0.75, 1, 1.5, 1.75\}.

\subsection{Evaluations on Cityscapes}
\paragraph{Dataset description.}
Cityscapes~\cite{Cityscapes} is a large urban street scene parsing benchmark.
It contains $2,975$ fine annotation images for training, $500$ images for validation, $1,525$ images for testing and extra $20,000$ coarsely annotated images for training.
We only use the fine annotation set in our experiments.
It includes $19$ categories for evaluation.

\begin{table}[t]
\centering
\small
\tablestyle{6pt}{1.05}
\begin{tabular}{l|c|l|x{22}}
\shline
\multicolumn{1}{c|}{model}  & reference & \multicolumn{1}{c|}{backbone} & \emph{mIoU} \\
\hline
RefineNet~\cite{Lin-CVPR-Refinenet-2017}  & CVPR2017 & ResNet-101 & 73.6 \\
GCN~\cite{Peng-CVPR-Largekernl-2017} 	  & CVPR2017 & ResNet-101 & 76.9 \\
DUC~\cite{Wang-WACV-DUC-2018} 		      & WACV2018 & ResNet-101 & 77.6 \\
DSSPN~\cite{Liang-DSSPN-CVPR-2018} 		  & CVPR2018 & ResNet-101 & 77.8 \\
SAC~\cite{Zhang-ICCV-SAC-2017} 		      & ICCV2017 & ResNet-101 & 78.1 \\
PSPNet~\cite{Zhao-CVPR-PSPNet-2017} 	  & CVPR2017 & ResNet-101 & 78.4 \\
BiSeNet~\cite{Yu-ECCV-BiSeNet-2018} 	  & ECCV2018 & ResNet-101 & 78.9 \\
AAF~\cite{Ke-ECCV-AAF-2018} 			  & ECCV2018 & ResNet-101 & 79.1 \\
DFN~\cite{Yu-CVPR-DFN-2018} 			  & CVPR2018 & ResNet-101 & 79.3 \\
PSANet~\cite{Zhao-ECCV-PSANet-2018} 	  & ECCV2018 & ResNet-101 & 80.1 \\
DenseASPP~\cite{Yang-CVPR-DenseASPP-2018} & CVPR2018 & DenseNet-161 & \underline{80.6} \\
ANL~\cite{Zhu-ICCV-ANL-2019} 			  & ICCV2019 & ResNet-101 & \textbf{81.3} \\
\hline
CPNet101 & - & ResNet-101 & \textbf{81.3}   \\
\shline
\end{tabular}
\vspace{0.5em}
\caption{
Quantitative evaluations on the Cityscapes test set. 
The proposed CPNet performs favorably against \emph{state-of-the-art} segmentation methods.
We only list the methods training with merely the fine dataset.
}
\label{tab:res-city}
\end{table}

\paragraph{Comparison with state-of-the-art.}
Table~\ref{tab:res-city} lists the performance results of other \emph{state-of-the-art} methods and our CPNet.
We adopt the multi-scale and flipped testing strategy on our experiments.
Following the pioneering work~\cite{Peng-CVPR-Largekernl-2017, Yu-CVPR-DFN-2018, Yu-ECCV-BiSeNet-2018}, we train our model with both the train-fine set and val-fine set to improve the performance on the test set.
Our CPNet achieves $81.3\%$ mIoU on the Cityscapes test set only with the fine dataset, which outperforms the DenseASPP based on DenseNet-161~\cite{Huang-CVPR-DenseNet-2017} by $0.9$ point.

\section{Concluding Remarks}
In this work, we construct an effective Context Prior for scene segmentation.
It distinguishes the different contextual dependencies with the supervision of the proposed Affinity Loss.
To embed the Context Prior into the network, we present a Context Prior Network, composed of a backbone network and a Context Prior Layer.
The Aggregation Module is applied to aggregate spatial information for reasoning the contextual relationship and embedded into the Context Prior Layer. 
Extensive quantitative and qualitative comparison shows that the proposed CPNet performs favorably against recent {state-of-the-art} scene segmentation approaches.

\section*{Acknowledgements}
This work is supported by the National Natural Science Foundation of China (No. 61433007 and 61876210).

\clearpage

{\small
\bibliographystyle{ieee_fullname}
\bibliography{reference}
}

\end{document}

%% file: macro.tex
\usepackage{times}
\usepackage{epsfig}
\usepackage{xspace}
\usepackage{color}
\usepackage{multirow}
\usepackage{graphics}
\usepackage{adjustbox}
\usepackage{subfigure}
\usepackage{amsmath}
\usepackage{algorithmicx,algorithm}
\usepackage{paralist} %
\usepackage{enumitem}
\usepackage{graphicx} %
\usepackage{epstopdf}%
\usepackage{booktabs} %

\usepackage[pagebackref=true,breaklinks=true,letterpaper=true,colorlinks,citecolor=blue,linkcolor=blue,bookmarks=false]{hyperref}

\long\def\ignorethis#1{}

\definecolor{Gray}{rgb}{0.35,0.35,0.35}
\definecolor{Blue}{rgb}{0,0.2,0.8}
\definecolor{Red}{rgb}{0.8,0.2,0}
\definecolor{Green}{rgb}{0.0,0.5,0.1}
\definecolor{Gray}{rgb}{0.4,0.4,0.4}

\newlength\paramargin
\newlength\figmargin
\newlength\secmargin

\usepackage{array}
\newcolumntype{L}[1]{>{\raggedright\let\newline\\\arraybackslash\hspace{0pt}}m{#1}}
\newcolumntype{C}[1]{>{\centering\let\newline\\\arraybackslash\hspace{0pt}}m{#1}}
\newcolumntype{R}[1]{>{\raggedleft\let\newline\\\arraybackslash\hspace{0pt}}m{#1}}

\def\ie{i.e.,~}
\def\eg{e.g.,~}

\graphicspath{{figure/eps/}}

%% file: main.bbl
\begin{thebibliography}{10}\itemsep=-1pt

\bibitem{Chen-ICLR-Deeplabv2-2016}
Liang-Chieh Chen, George Papandreou, Iasonas Kokkinos, Kevin Murphy, and Alan~L
  Yuille.
\newblock Semantic image segmentation with deep convolutional nets and fully
  connected crfs.
\newblock {\em Proc. International Conference on Learning Representations
  (ICLR)}, 2015.

\bibitem{Chen-Arxiv-Deeplabv2-2016}
Liang-Chieh Chen, George Papandreou, Iasonas Kokkinos, Kevin Murphy, and Alan~L
  Yuille.
\newblock Deeplab: Semantic image segmentation with deep convolutional nets,
  atrous convolution, and fully connected crfs.
\newblock {\em arXiv}, 2016.

\bibitem{Chen-Arxiv-Deeplabv3-2017}
Liang-Chieh Chen, George Papandreou, Florian Schroff, and Hartwig Adam.
\newblock Rethinking atrous convolution for semantic image segmentation.
\newblock {\em arXiv}, 2017.

\bibitem{Chen-CVPR-AttentionScale-2016}
Liang-Chieh Chen, Yi Yang, Jiang Wang, Wei Xu, and Alan~L Yuille.
\newblock Attention to scale: Scale-aware semantic image segmentation.
\newblock In {\em Proc. IEEE Conference on Computer Vision and Pattern
  Recognition (CVPR)}, 2016.

\bibitem{Chen-ECCV-Deeplabv3p-2018}
Liang-Chieh Chen, Yukun Zhu, George Papandreou, Florian Schroff, and Hartwig
  Adam.
\newblock Encoder-decoder with atrous separable convolution for semantic image
  segmentation.
\newblock In {\em Proc. European Conference on Computer Vision (ECCV)}, pages
  801--818, 2018.

\bibitem{Chen-NIPS-A2Net-2018}
Yunpeng Chen, Yannis Kalantidis, Jianshu Li, Shuicheng Yan, and Jiashi Feng.
\newblock A\^{} 2-nets: Double attention networks.
\newblock In {\em Proc. Advances in Neural Information Processing Systems
  (NeurIPS)}, pages 352--361, 2018.

\bibitem{Chollet-CVPR-Xception-2017}
Fran{\c{c}}ois Chollet.
\newblock Xception: Deep learning with depthwise separable convolutions.
\newblock 2017.

\bibitem{Cityscapes}
Marius Cordts, Mohamed Omran, Sebastian Ramos, Timo Rehfeld, Markus Enzweiler,
  Rodrigo Benenson, Uwe Franke, Stefan Roth, and Bernt Schiele.
\newblock The cityscapes dataset for semantic urban scene understanding.
\newblock In {\em Proc. IEEE Conference on Computer Vision and Pattern
  Recognition (CVPR)}, 2016.

\bibitem{Dai-ICCV-BoxSup-2015}
Jifeng Dai, Kaiming He, and Jian Sun.
\newblock Boxsup: Exploiting bounding boxes to supervise convolutional networks
  for semantic segmentation.
\newblock In {\em Proc. IEEE International Conference on Computer Vision
  (ICCV)}, pages 1635--1643, 2015.

\bibitem{Ding-CVPR-CCL-2018}
Henghui Ding, Xudong Jiang, Bing Shuai, Ai Qun~Liu, and Gang Wang.
\newblock Context contrasted feature and gated multi-scale aggregation for
  scene segmentation.
\newblock In {\em Proc. IEEE Conference on Computer Vision and Pattern
  Recognition (CVPR)}, pages 2393--2402, 2018.

\bibitem{Fu-CVPR-DANet-2019}
Jun Fu, Jing Liu, Haijie Tian, Zhiwei Fang, and Hanqing Lu.
\newblock Dual attention network for scene segmentation.
\newblock {\em Proc. IEEE Conference on Computer Vision and Pattern Recognition
  (CVPR)}, 2019.

\bibitem{Kitti}
Andreas Geiger, Philip Lenz, and Raquel Urtasun.
\newblock Are we ready for autonomous driving? the kitti vision benchmark
  suite.
\newblock In {\em Proc. IEEE Conference on Computer Vision and Pattern
  Recognition (CVPR)}, pages 3354--3361, 2012.

\bibitem{He-CVPR-ResNet-2016}
Kaiming He, Xiangyu Zhang, Shaoqing Ren, and Jian Sun.
\newblock Deep residual learning for image recognition.
\newblock In {\em Proc. IEEE Conference on Computer Vision and Pattern
  Recognition (CVPR)}, 2016.

\bibitem{Howard-Arxiv-MobileNet-2017}
Andrew~G Howard, Menglong Zhu, Bo Chen, Dmitry Kalenichenko, Weijun Wang,
  Tobias Weyand, Marco Andreetto, and Hartwig Adam.
\newblock Mobilenets: Efficient convolutional neural networks for mobile vision
  applications.
\newblock {\em arXiv}, 2017.

\bibitem{Hu-CVPR-RelationNet-2018}
Han Hu, Jiayuan Gu, Zheng Zhang, Jifeng Dai, and Yichen Wei.
\newblock Relation networks for object detection.
\newblock In {\em Proc. IEEE Conference on Computer Vision and Pattern
  Recognition (CVPR)}, 2018.

\bibitem{Hu-CVPR-SEnet-2017}
Jie Hu, Li Shen, and Gang Sun.
\newblock Squeeze-and-excitation networks.
\newblock {\em Proc. IEEE Conference on Computer Vision and Pattern Recognition
  (CVPR)}, 2018.

\bibitem{Huang-CVPR-DenseNet-2017}
Gao Huang, Zhuang Liu, Laurens van~der Maaten, and Kilian~Q. Weinberger.
\newblock Densely connected convolutional networks.
\newblock {\em Proc. IEEE Conference on Computer Vision and Pattern Recognition
  (CVPR)}, pages 2261--2269, 2017.

\bibitem{Huang-ICCV-CCNet-2018}
Zilong Huang, Xinggang Wang, Chang Huang, Yunchao Wei, and Wenyu Liu.
\newblock Ccnet: Criss-cross attention for semantic segmentation.
\newblock {\em Proc. IEEE International Conference on Computer Vision (ICCV)},
  2019.

\bibitem{Hung-ICCV-GCPNet-2017}
Wei-Chih Hung, Yi-Hsuan Tsai, Xiaohui Shen, Zhe Lin, Kalyan Sunkavalli, Xin Lu,
  and Ming-Hsuan Yang.
\newblock Scene parsing with global context embedding.
\newblock In {\em Proc. IEEE International Conference on Computer Vision
  (ICCV)}, 2017.

\bibitem{Ioffe-ICML-BN-2015}
Sergey Ioffe and Christian Szegedy.
\newblock Batch normalization: Accelerating deep network training by reducing
  internal covariate shift.
\newblock In {\em Proc. International Conference on Machine Learning (ICML)},
  pages 448--456, 2015.

\bibitem{Ke-ECCV-AAF-2018}
Tsung-Wei Ke, Jyh-Jing Hwang, Ziwei Liu, and Stella~X Yu.
\newblock Adaptive affinity fields for semantic segmentation.
\newblock In {\em Proc. European Conference on Computer Vision (ECCV)}, pages
  587--602, 2018.

\bibitem{Krizhevsky-NIPS-Imagenet}
Alex Krizhevsky, Ilya Sutskever, and Geoffrey~E Hinton.
\newblock Imagenet classification with deep convolutional neural networks.
\newblock In {\em Proc. Advances in Neural Information Processing Systems
  (NeurIPS)}, 2012.

\bibitem{Li-BMVC-PAN-2018}
Hanchao Li, Pengfei Xiong, Jie An, and Lingxue Wang.
\newblock Pyramid attention network for semantic segmentation.
\newblock {\em Proc. the British Machine Vision Conference (BMVC)}, 2018.

\bibitem{Liang-DSSPN-CVPR-2018}
Xiaodan Liang, Hongfei Zhou, and Eric~P. Xing.
\newblock Dynamic-structured semantic propagation network.
\newblock {\em Proc. IEEE Conference on Computer Vision and Pattern Recognition
  (CVPR)}, pages 752--761, 2018.

\bibitem{Lin-CVPR-Refinenet-2017}
Guosheng Lin, Anton Milan, Chunhua Shen, and Ian Reid.
\newblock Refinenet: Multi-path refinement networks with identity mappings for
  high-resolution semantic segmentation.
\newblock {\em Proc. IEEE Conference on Computer Vision and Pattern Recognition
  (CVPR)}, 2017.

\bibitem{Liu-CVPR-OANet-2019}
Huanyu Liu, Chao Peng, Changqian Yu, Jingbo Wang, Xu Liu, Gang Yu, and Wei
  Jiang.
\newblock An end-to-end network for panoptic segmentation.
\newblock {\em Proc. IEEE Conference on Computer Vision and Pattern Recognition
  (CVPR)}, pages 6165--6174, 2019.

\bibitem{Liu-ARXIV-ParseNet-2016}
Wei Liu, Andrew Rabinovich, and Alexander~C Berg.
\newblock Parsenet: Looking wider to see better.
\newblock {\em arXiv}, 2016.

\bibitem{Long-CVPR-FCN-2015}
Jonathan Long, Evan Shelhamer, and Trevor Darrell.
\newblock Fully convolutional networks for semantic segmentation.
\newblock In {\em Proc. IEEE Conference on Computer Vision and Pattern
  Recognition (CVPR)}, 2015.

\bibitem{Ma-ECCV-Shufflenetv2-2018}
Ningning Ma, Xiangyu Zhang, Hai-Tao Zheng, and Jian Sun.
\newblock Shufflenet v2: Practical guidelines for efficient cnn architecture
  design.
\newblock In {\em Proc. European Conference on Computer Vision (ECCV)}, pages
  116--131, 2018.

\bibitem{PASCAL-Context}
Roozbeh Mottaghi, Xianjie Chen, Xiaobai Liu, Nam-Gyu Cho, Seong-Whan Lee, Sanja
  Fidler, Raquel Urtasun, and Alan Yuille.
\newblock The role of context for object detection and semantic segmentation in
  the wild.
\newblock In {\em Proc. IEEE Conference on Computer Vision and Pattern
  Recognition (CVPR)}, 2014.

\bibitem{PyTorch}
Adam Paszke, Sam Gross, Soumith Chintala, Gregory Chanan, Edward Yang, Zachary
  DeVito, Zeming Lin, Alban Desmaison, Luca Antiga, and Adam Lerer.
\newblock Automatic differentiation in pytorch.
\newblock In {\em Neural Information Processing Systems Workshops}, 2017.

\bibitem{Peng-CVPR-Largekernl-2017}
Chao Peng, Xiangyu Zhang, Gang Yu, Guiming Luo, and Jian Sun.
\newblock Large kernel matters--improve semantic segmentation by global
  convolutional network.
\newblock {\em Proc. IEEE Conference on Computer Vision and Pattern Recognition
  (CVPR)}, 2017.

\bibitem{Szegedy-CVPR-Inception-2016}
Christian Szegedy, Vincent Vanhoucke, Sergey Ioffe, Jonathon Shlens, and
  Zbigniew Wojna.
\newblock Rethinking the inception architecture for computer vision.
\newblock In {\em Proc. IEEE Conference on Computer Vision and Pattern
  Recognition (CVPR)}, pages 2818--2826, 2016.

\bibitem{Ashish-NIPS-Transfomer-2017}
Ashish Vaswani, Noam Shazeer, Niki Parmar, Jakob Uszkoreit, Llion Jones,
  Aidan~N. Gomez, Lukasz Kaiser, and Illia Polosukhin.
\newblock Attention is all you need.
\newblock In {\em Proc. Advances in Neural Information Processing Systems
  (NeurIPS)}, 2017.

\bibitem{Wang-TPAMI-HRNet-2019}
Jingdong Wang, Ke Sun, Tianheng Cheng, Borui Jiang, Chaorui Deng, Yang Zhao,
  Dong Liu, Yadong Mu, Mingkui Tan, Xinggang Wang, Wenyu Liu, and Bin Xiao.
\newblock Deep high-resolution representation learning for visual recognition.
\newblock 2019.

\bibitem{Wang-WACV-DUC-2018}
Panqu Wang, Pengfei Chen, Ye Yuan, Ding Liu, Zehua Huang, Xiaodi Hou, and
  Garrison Cottrell.
\newblock Understanding convolution for semantic segmentation.
\newblock {\em Proc. IEEE Winter Conference on Applications of Computer Vision
  (WACV)}, 2018.

\bibitem{Wang-CVPR-Nonlocal-2018}
Xiaolong Wang, Ross Girshick, Abhinav Gupta, and Kaiming He.
\newblock Non-local neural networks.
\newblock {\em Proc. IEEE Conference on Computer Vision and Pattern Recognition
  (CVPR)}, 2018.

\bibitem{Wu-Arxiv-HighPerf-2016}
Zifeng Wu, Chunhua Shen, and Anton van~den Hengel.
\newblock High-performance semantic segmentation using very deep fully
  convolutional networks.
\newblock {\em arXiv}, 2016.

\bibitem{Xiao-UperNet-ECCV-2018}
Tete Xiao, Yingcheng Liu, Bolei Zhou, Yuning Jiang, and Jian Sun.
\newblock Unified perceptual parsing for scene understanding.
\newblock In {\em Proc. European Conference on Computer Vision (ECCV)}, pages
  418--434, 2018.

\bibitem{Yang-CVPR-DenseASPP-2018}
Maoke Yang, Kun Yu, Chi Zhang, Zhiwei Li, and Kuiyuan Yang.
\newblock Denseaspp for semantic segmentation in street scenes.
\newblock In {\em Proc. IEEE Conference on Computer Vision and Pattern
  Recognition (CVPR)}, pages 3684--3692, 2018.

\bibitem{Yin-ICCV-VN-2019}
Wei Yin, Yifan Liu, Chunhua Shen, and Youliang Yan.
\newblock Enforcing geometric constraints of virtual normal for depth
  prediction.
\newblock {\em Proc. IEEE International Conference on Computer Vision (ICCV)},
  pages 5683--5692, 2019.

\bibitem{Yu-ECCV-BiSeNet-2018}
Changqian Yu, Jingbo Wang, Chao Peng, Changxin Gao, Gang Yu, and Nong Sang.
\newblock Bisenet: Bilateral segmentation network for real-time semantic
  segmentation.
\newblock In {\em Proc. European Conference on Computer Vision (ECCV)}, pages
  325--341, 2018.

\bibitem{Yu-CVPR-DFN-2018}
Changqian Yu, Jingbo Wang, Chao Peng, Changxin Gao, Gang Yu, and Nong Sang.
\newblock Learning a discriminative feature network for semantic segmentation.
\newblock In {\em Proc. IEEE Conference on Computer Vision and Pattern
  Recognition (CVPR)}, 2018.

\bibitem{Yuan-Arxiv-OCNet-2018}
Yuhui Yuan and Jingdong Wang.
\newblock Ocnet: Object context network for scene parsing.
\newblock {\em arXiv}, 2018.

\bibitem{Zhang-CVPR-EncNet-2018}
Hang Zhang, Kristin Dana, Jianping Shi, Zhongyue Zhang, Xiaogang Wang, Ambrish
  Tyagi, and Amit Agrawal.
\newblock Context encoding for semantic segmentation.
\newblock In {\em Proc. IEEE Conference on Computer Vision and Pattern
  Recognition (CVPR)}, pages 7151--7160, 2018.

\bibitem{Zhang-CVPR-ACF-2019}
Hang Zhang, Han Zhang, Chenguang Wang, and Junyuan Xie.
\newblock Co-occurrent features in semantic segmentation.
\newblock In {\em Proc. IEEE Conference on Computer Vision and Pattern
  Recognition (CVPR)}, pages 548--557, 2019.

\bibitem{Zhang-ICCV-SAC-2017}
Rui Zhang, Sheng Tang, Yongdong Zhang, Jintao Li, and Shuicheng Yan.
\newblock Scale-adaptive convolutions for scene parsing.
\newblock In {\em Proc. IEEE International Conference on Computer Vision
  (ICCV)}, pages 2031--2039, 2017.

\bibitem{Zhang-CVPR-Shufflenet-2018}
Xiangyu Zhang, Xinyu Zhou, Mengxiao Lin, and Jian Sun.
\newblock Shufflenet: An extremely efficient convolutional neural network for
  mobile devices.
\newblock In {\em Proc. IEEE Conference on Computer Vision and Pattern
  Recognition (CVPR)}, pages 6848--6856, 2018.

\bibitem{Zhao-CVPR-PSPNet-2017}
Hengshuang Zhao, Jianping Shi, Xiaojuan Qi, Xiaogang Wang, and Jiaya Jia.
\newblock Pyramid scene parsing network.
\newblock {\em Proc. IEEE Conference on Computer Vision and Pattern Recognition
  (CVPR)}, 2017.

\bibitem{Zhao-ECCV-PSANet-2018}
Hengshuang Zhao, Yi Zhang, Shu Liu, Jianping Shi, Chen~Change Loy, Dahua Lin,
  and Jiaya Jia.
\newblock {PSANet}: Point-wise spatial attention network for scene parsing.
\newblock In {\em Proc. European Conference on Computer Vision (ECCV)}, 2018.

\bibitem{Zheng-ICCV-CRFasRNN-2015}
Shuai Zheng, Sadeep Jayasumana, Bernardino Romera-Paredes, Vibhav Vineet,
  Zhizhong Su, Dalong Du, Chang Huang, and Philip~HS Torr.
\newblock Conditional random fields as recurrent neural networks.
\newblock In {\em Proc. IEEE International Conference on Computer Vision
  (ICCV)}, 2015.

\bibitem{Zhou-ADE-2017}
Bolei Zhou, Hang Zhao, Xavier Puig, Sanja Fidler, Adela Barriuso, and Antonio
  Torralba.
\newblock Scene parsing through ade20k dataset.
\newblock In {\em Proc. IEEE Conference on Computer Vision and Pattern
  Recognition (CVPR)}, 2017.

\bibitem{Zhu-ICCV-ANL-2019}
Zhen Zhu, Mengde Xu, Song Bai, Tengteng Huang, and Xiang Bai.
\newblock Asymmetric non-local neural networks for semantic segmentation.
\newblock In {\em Proc. IEEE International Conference on Computer Vision
  (ICCV)}, pages 593--602, 2019.

\end{thebibliography}
